\documentclass[lettersize,journal]{IEEEtran}

\IEEEoverridecommandlockouts
\usepackage{cite}
\usepackage[caption=false,font=normalsize,labelfont=sf,textfont=sf]{subfig}
\usepackage{amsmath,amssymb,amsfonts}
\usepackage{algorithmic}
\usepackage{graphicx}
\usepackage{textcomp}
\usepackage{xcolor}
\usepackage{booktabs, longtable}
\usepackage{enumitem,booktabs,threeparttable}
\usepackage[most]{tcolorbox}
\tcbuselibrary{listings, breakable}
\usepackage{stfloats}
\usepackage{enumitem}
\usepackage{amsmath}
\usepackage{threeparttable}
\usepackage{multirow}
\usepackage{newtxtext}
\usepackage{hyperref}
\usepackage{graphicx}
\usepackage{subcaption}
\usepackage{float} 
\usepackage{caption}
\usepackage{array}
\usepackage{etoolbox} 
\usepackage{placeins}

\def\BibTeX{{\rm B\kern-.05em{\sc i\kern-.025em b}\kern-.08em
    T\kern-.1667em\lower.7ex\hbox{E}\kern-.125emX}}
\begin{document}

\title{DK-Root: A Joint Data-and-Knowledge-Driven Framework for Root Cause Analysis of QoE Degradations in Mobile Networks}

\author{Qizhe Li, Haolong Chen, Jiansheng Li, Shuqi Chai, \\
Xuan Li, Yuzhou Hou, Xinhua Shao, Fangfang Li, Kaifeng Han, Guangxu Zhu,~\IEEEmembership{Member,~IEEE}%
\thanks{Qizhe Li, Haolong Chen, Jiansheng Li, Shuqi Chai and Guangxu Zhu are with Shenzhen Research Institute of Big Data, Shenzhen, China, and the School of Science and Engineering, The Chinese University of Hong Kong, Shenzhen, China (Emails: \{qizheli, haolongchen1, jianshengli\}@link.cuhk.edu.cn, \{schai, gxzhu\}@sribd.cn).}%
\thanks{Xuan Li and Yuzhou Hou are with the Experience Lab, Huawei Technologies Co., Ltd., Shenzhen, China (Emails: \{lixuan69, houyuzhou\}@huawei.com).}%
\thanks{Xinhua Shao and Fangfang Li are with China Telecommunications Group Co., Ltd., Beijing, China (Email: \{shaoxh, liff\}@chinatelecom.cn).}%
\thanks{Kaifeng Han is with the China Academy of Information and Communications Technology, Beijing, China (Email: hankaifeng@caict.ac.cn).}%
\thanks{Corresponding author: Guangxu Zhu (gxzhu@sribd.cn).}%
}

\markboth{Journal of \LaTeX\ Class Files,~Vol.~14, No.~8, August~2021}%
{Shell \MakeLowercase{\textit{et al.}}: A Sample Article Using IEEEtran.cls for IEEE Journals}

\IEEEpubid{0000--0000/00\$00.00~\copyright~2021 IEEE}

\maketitle

\begin{abstract}
Diagnosing the root causes of Quality of Experience (QoE) degradations in operational mobile networks is challenging due to complex cross-layer interactions among kernel performance indicators (KPIs) and the scarcity of reliable expert annotations. Although rule-based heuristics can generate labels at scale, they are noisy and coarse-grained, limiting the accuracy of purely data-driven approaches. To address this, we propose DK-Root, a joint data–and-knowledge-driven framework that unifies scalable weak supervision with precise expert guidance for robust root-cause analysis. DK-Root first pretrains an encoder via contrastive representation learning using abundant rule-based labels while explicitly denoising their noise through a supervised contrastive objective. To supply task-faithful data augmentation, we introduce a class-conditional diffusion model that generates KPIs sequences preserving root-cause semantics, and by controlling reverse diffusion steps, it produces weak and strong augmentations that improve intra-class compactness and inter-class separability. Finally, the encoder and the lightweight classifier are jointly fine-tuned with scarce expert-verified labels to sharpen decision boundaries. Extensive experiments on a real-world, operator-grade dataset demonstrate state-of-the-art accuracy, with DK-Root surpassing traditional ML and recent semi-supervised time-series methods. Ablations confirm the necessity of the conditional diffusion augmentation and the pretrain–finetune design, validating both representation quality and classification gains.
\end{abstract}


\section{Introduction}
\begin{figure}
    \centering
    \includegraphics[width=\linewidth]{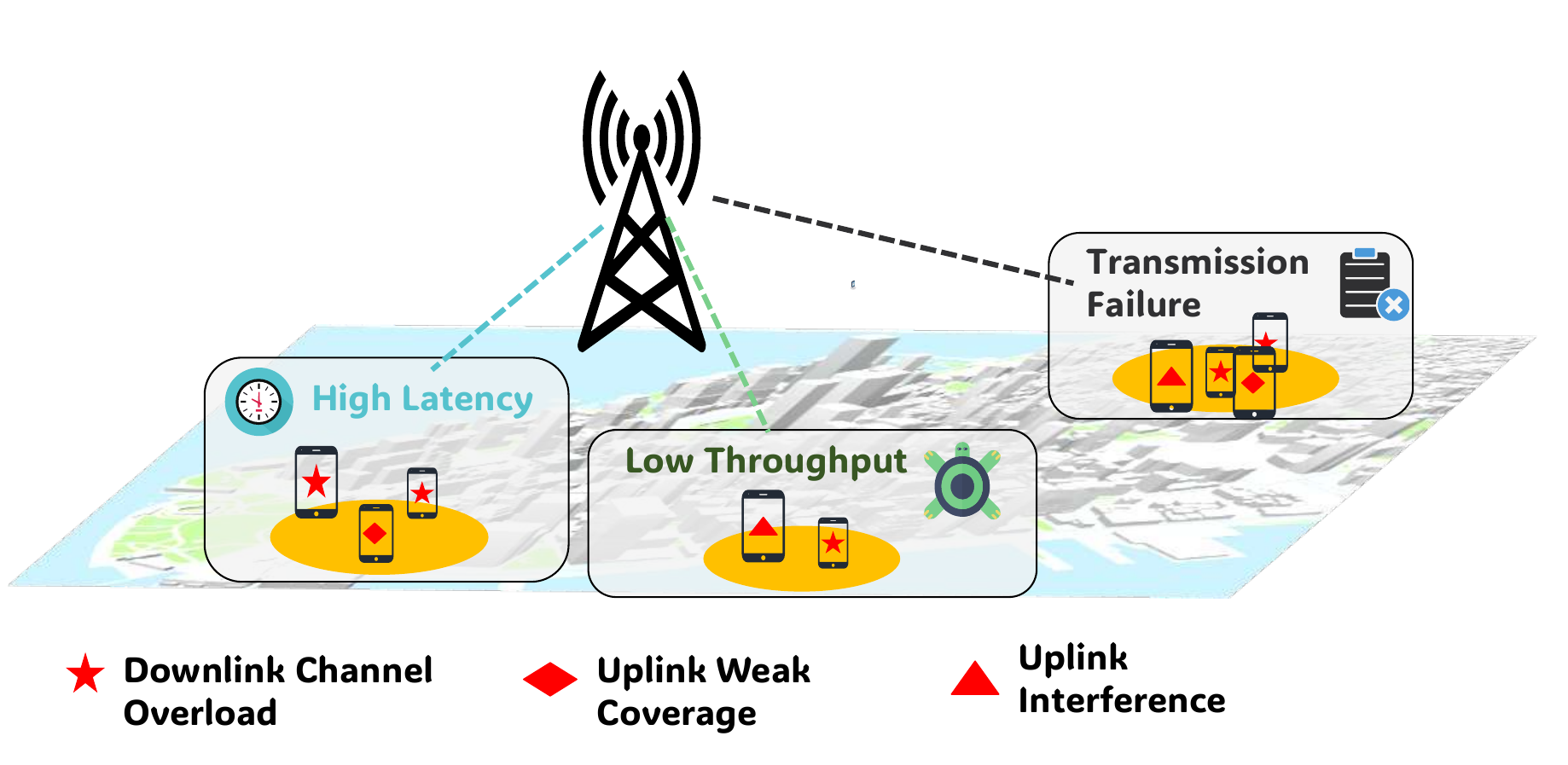}
    \caption{Poor Communication Quality of Experience Caused by Different Root-causes.}
    \label{fig:background}
\end{figure}

\IEEEPARstart{E}{nsuring} high Quality of Experience (QoE) in mobile networks is increasingly difficult as architectures grow more complex, with tightly coupled protocol layers and heterogeneous elements. These cross-layer dependencies obscure how degradations arise and propagate, slowing diagnosis and remediation. At the same time, the push toward 6G—with its vision of autonomous operation and embedded intelligence—magnifies the need for reliable, data-driven root-cause analysis. Recent directions such as statistical digital twins~\cite{luo2023srcon} seek closed-loop integration between physical infrastructure and virtual models for real-time inference and adaptation. Within this context, accurately identifying the root causes of QoE drops (e.g., excessive latency or low throughput) from multivariate KPIs time series is pivotal. As illustrated in Figure~\ref{fig:background}, degradations can stem from diverse factors including interference, weak coverage, or resource congestion, and their signatures often span the physical (PHY), medium access control (MAC), radio link control (RLC), and packet data convergence protocol (PDCP) layers. However, constructing robust diagnostic models remains challenging due to scarce, costly expert annotations and the brittleness of rule-based heuristics, which scale labeling but introduce noise and coarse granularity. These realities call for frameworks that can leverage abundant weak supervision without sacrificing semantic fidelity, while selectively incorporating expert knowledge to refine decision boundaries.

\IEEEpubidadjcol

Existing methods for QoE root-cause analysis broadly fall into two categories: rule-based and learning-based approaches. Rule-based methods encode domain knowledge into handcrafted logic, such as threshold conditions on signal-to-interference-plus-noise ratio (SINR) or reference signal received power (RSRP), and are widely used due to their simplicity~\cite{ciocarlie2014feasibility, moulay2020novel, yuan2020anomaly}. However, they tend to be brittle when facing subtle, compound, or previously unseen degradation patterns. In contrast, deep learning methods have demonstrated the ability to extract complex cross-layer features from KPIs streams~\cite{chawla2020interpretable, sun2024spotlight}. Yet, their performance is strongly dependent on the availability of high-quality annotated training data—a requirement that is difficult to meet in real-world network environments.

A key bottleneck in mobile network diagnosis lies in the scarcity of reliable root-cause annotations. Each communication record comprises multivariate KPIs sequences spanning multiple protocol layers (e.g., PHY, MAC, RLC, PDCP), whose semantics are neither intuitive nor visually interpretable. Unlike image or audio data, understanding these KPIs sequences—and more importantly, their temporal and cross-layer interactions—requires deep domain expertise in mobile protocol behaviors and fault propagation \cite{chen2024overview}. As a result, expert-verified root-cause labels are both expensive and limited in scale.
To scale up supervision, practitioners often resort to rule-based heuristics derived from KPIs thresholds and logical expressions to generate large quantities of weak labels. However, these rule-based labels tend to be coarse-grained and noisy due to their simplistic assumptions and lack of context awareness. This leads to a significant label quality gap between abundant but imprecise rule labels and accurate yet scarce expert annotations.

To bridge this gap, semi-supervised learning (SSL) has gained increasing traction. In particular, semi-supervised contrastive learning (SSCL) has emerged as a powerful paradigm for time-series tasks~\cite{zhang2024self}, as it enables discriminative representation learning using a small number of high-quality labels. By encouraging samples with similar degradation patterns to cluster together (positive pairs) while pushing dissimilar ones apart (negative pairs), SSCL facilitates robust feature learning under weak supervision.

However, the effectiveness of SSCL heavily relies on the design of data augmentations. In the time-series domain, augmentation-based contrastive learning is commonly used~\cite{wen2020time, peng2021fault, luo2023time}, with transformations such as jittering, scaling, or time-warping applied to generate diverse views of the same sequence. Yet, these generic augmentations are often task-agnostic and risk violating the physical or protocol-layer constraints embedded in KPIs. For example, injecting noise into RSRP or shuffling SINR sequences may disrupt signal continuity or scheduling semantics, leading to semantic drift and degraded diagnostic performance.

These challenges motivate the need for communication-aware and cause-consistent data augmentation strategies that preserve the structural integrity and semantic patterns of KPIs sequences. In particular, generative models that can synthesize augmented views conditioned on specific degradation types offer a promising direction, as they enable the creation of diverse yet semantically aligned samples for contrastive training.

To address the challenge of root-cause analysis for QoE degradations under limited supervision, we propose \textbf{DK-Root}, a joint data-and-knowledge-driven framework tailored for this task in mobile networks. DK-Root is motivated by the observation that large-scale rule-based root-cause annotations—representing a data-driven paradigm—are inevitably noisy, whereas expert-verified diagnostic results—reflecting knowledge-driven supervision—are sparse but reliable. To effectively leverage both sources, we unify coarse-grained representation learning and fine-grained expert refinement into a single framework. Our key contributions are as follows:

\begin{itemize}
    \item \textbf{A Joint Data-and-Knowledge QoE Root-Cause Analysis Framework:} We design DK-Root to unify two complementary supervision sources: (i) rule-based labels that enable scalable but coarse data-driven supervision, and (ii) expert-verified labels that provide precise but scarce knowledge-driven guidance. To bridges the gap between data abundance and knowledge scarcity, we decouple feature representation learning and final classification, allowing the model to benefit from both broad heuristics and high-quality human expertise.

    \item \textbf{Conditional Generative Augmentation under Expert Guidance:} We introduce a conditional diffusion-based generative module that synthesizes class-consistent KPIs sequences. Expert labels are first embedded and fused with KPIs representations through a cross-modal fusion network, forming semantically aligned conditional inputs. We further control the reverse diffusion time step to produce augmented samples of varying semantic intensities: weak augmentation (small time steps) ensures stability, while strong augmentation (large time steps) introduces higher diversity. These expert-conditioned augmentations improve intra-class compactness and enhance generalization under limited supervision.

    \item \textbf{Data-Aware Contrastive Representation Learning with Knowledge-Based Refinement:} We employ a supervised contrastive learning strategy to construct a degradation-aware embedding space. By pulling together samples with similar rule-based labels and pushing apart those with different root causes, the model learns to denoise noisy supervision while capturing the underlying structure of degradation patterns. This contrastive pretraining is followed by expert-guided fine-tuning using a small number of high-fidelity labels, which further refines the classification boundary and improves root-cause discrimination in noisy and low-resource settings.

    \item \textbf{Comprehensive Validation in Realistic Settings:} We evaluate the proposed DK-Root framework on real-world mobile network datasets annotated with both rule-based and expert labels. Through extensive experiments—including comparisons with multiple machine learning and deep learning baselines, qualitative assessment of the generated samples, and ablation studies—we demonstrate consistent improvements in QoE root-cause analysis accuracy and superiority.
\end{itemize}

The remainder of this paper is organized as follows. Section~\ref{sec:related_work} reviews related work. Section~\ref{task description} defines the problem and notation. Section~\ref{sec:proposed method} presents our proposed framework. Section~\ref{sec:experiments} reports experimental results and analysis. Section~\ref{conclusion} concludes the paper and discusses future work.

\section{Related Work}
\label{sec:related_work}

\subsection{QoE Anomaly Detection}

QoE root-cause analysis has emerged as a key research topic in mobile communication networks, driven by the growing need for automated and intelligent network operation. Existing approaches can be broadly divided into two categories: traditional rule-based methods and modern data-driven learning models.

\noindent\textbf{Traditional Rule-Based Methods.}
Early works focused on encoding expert knowledge into static rule sets or statistical models for anomaly detection and root-cause inference. Ciocarlie et al.~\cite{ciocarlie2014feasibility} examined the feasibility of deploying anomaly detection systems within operational cellular networks, proposing probabilistic rule-based frameworks that rely on Markov logic networks. Their approach incorporated domain heuristics to detect network malfunctions, but required manual rule design and struggled with generalization across cell types and deployment contexts. Moulay et al.~\cite{moulay2020novel} proposed a decision-tree-based system for automated detection and classification of networking anomalies. While computationally efficient, their method is sensitive to pre-defined thresholding logic and suffers from rigidity in the face of unseen faults or cross-layer interactions. Yuan et al.~\cite{yuan2020anomaly} explored the integration of AI into the root-cause analysis workflow. They proposed a hybrid anomaly detection system combining statistical profiling and shallow machine learning, with a focus on deployment viability in large-scale radio networks. However, their approach still heavily relies on manually defined KPIs patterns, limiting scalability. Ramírez et al.~\cite{ramirez2023explainable} contributed to the explainability of anomaly detection by incorporating interpretable machine learning models (e.g., decision trees and SHAP values) for mobile network performance monitoring. While improving transparency, the method inherits the generalization limitations of rule-centric systems and cannot fully leverage unlabeled data.

\noindent\textbf{Deep Learning and Unsupervised Models.}
More recently, researchers have turned to deep learning to exploit the rich structure in time-series KPIs data. Chawla et al.~\cite{chawla2020interpretable} proposed an interpretable unsupervised anomaly detection method for RAN cell traces, using autoencoder-based architectures to uncover latent anomalies in network traces. Their framework emphasized explainability through feature attribution, but its unsupervised nature limits precise root-cause classification. Sun et al.~\cite{sun2024spotlight} developed SpotLight, a high-accuracy, explainable, and efficient anomaly detection system tailored for Open RAN environments. The model leverages hierarchical spatial-temporal feature extraction combined with outlier scoring to detect performance drops at scale. However, although SpotLight offers strong anomaly detection capabilities, it stops short of fine-grained root-cause categorization, leaving the diagnosis problem open.

Despite recent advances, existing approaches face inherent trade-offs: rule-based methods offer interpretability but lack flexibility, while deep learning models require large volumes of high-quality labels that are costly to obtain. Moreover, most prior work overlooks the integration of protocol-specific constraints, risking semantic distortion of critical KPIs (e.g., RSRP, SINR) during training or augmentation. In contrast, our work introduces a domain-aware semi-supervised contrastive learning framework that explicitly incorporates communication domain knowledge. By combining expert-labeled and rule-labeled data, and embedding protocol semantics into the learning process, our method balances label efficiency with model generalizability—bridging the gap between rigid rule systems and data-hungry deep models.

\subsection{Time-series Classification}

\noindent\textbf{Traditional Methods.}  
Classic TSC methods often rely on distance-based comparisons such as 1-Nearest Neighbor with Dynamic Time Warping (DTW), which remain strong baselines~\cite{bagnall2017great}. Enhancements like shapelets or symbolic representations can boost performance without extensive learning~\cite{baydogan2015learning}. These methods offer interpretability and simplicity, but struggle when handling high-dimensional, multimodal sequences and lack the ability to learn temporal abstractions.

\noindent\textbf{Deep Learning for TSC.}  
The rise of deep learning has significantly improved TSC performance. Discriminative models based on 1D-CNNs, RNNs (LSTM/GRU), and more recently Transformer architectures now lead performance benchmarks~\cite{ismail2019deep, ismail2020inceptiontime}. Noteworthy is the InceptionTime ensemble, which rivals or surpasses traditional ensembles like HIVE-COTE in both accuracy and scalability~\cite{ismail2020inceptiontime}. Reviews consistently affirm that deep models outperform conventional methods in the majority of benchmarked tasks~\cite{ismail2019deep, fawaz2020deep}. However, these models demand large training sets and often fail to generalize when labeled data are scarce. To address label scarcity, semi-supervised and self-supervised methods have emerged. Contrastive learning frameworks—e.g., TS-TCC (Temporal and Contextual Contrast) and its extension CA-TCC (Class-Aware TS-TCC)—achieve strong representation quality by enforcing consistency across augmented views of the same sequence~\cite{eldele2021time, yue2022ts2vec, eldele2023self}. TS2Vec introduces hierarchical contrastive objectives to learn embeddings at multiple temporal scales~\cite{yue2022ts2vec}. While powerful, these methods typically employ generic augmentations such as jittering, scaling, or warping. Other semi-supervised strategies adapted from computer vision, which integrate pseudo-labeling or self-training, also show improvement when paired with task-specific architectures and augmentations. Auxiliary self-supervised tasks like reconstruction or forecasting further enhance robustness under low-label conditions~\cite{eldele2023self}. 

Despite these advances, most semi-supervised TSC methods remain domain-agnostic and overlook signal-specific constraints. Augmentations that distort amplitude or time structure can corrupt the semantics of physical or protocol-driven signals—such as network KPIs (e.g., RSRP, SINR)—and adversely affect downstream classification. Building on these insights, we propose a domain-aware semi-supervised contrastive learning framework tailored for QoE root-cause analysis. By integrating expert knowledge into augmentation design and contrastive objectives, we ensure that learned embeddings maintain the rich temporal and protocol-layer semantics of network KPIs sequences. Moreover, our unified learning pipeline fuses expert-labeled and rule-labeled samples, delivering a balance of label efficiency, precision, and robustness.

\section{Problem Formulation}
\label{task description}

\begin{figure*}[!t]
    \centering
    \includegraphics[width=0.9\linewidth]{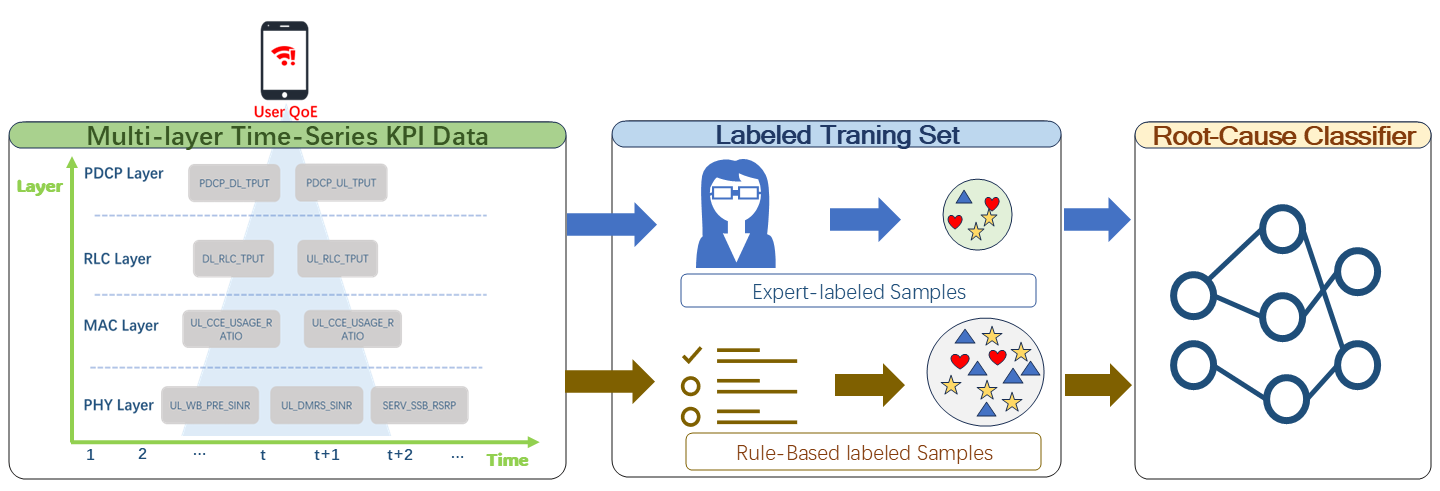}
    \caption{Overview of the root-cause analysis task for QoE degradations in mobile networks.}
    \label{fig:bigpicture}
\end{figure*}

We formalize the root-cause analysis of QoE degradations in mobile networks as a classification learning problem over multi-dimensional KPIs time-series data, where each sample corresponds to a sequence of communication performance metrics collected across multiple protocol layers. 

In this paper, we focus on a representative case of high-latency scenarios, aiming to accurately classify the underlying root causes based on KPIs patterns across different protocol layers.
High latency can result from various factors. In our study, we consider six common and representative root causes. The associated root-cause label $y \in {1, 2, \ldots, 6}$ corresponds to one of six degradation categories, covering typical issues such as interference, weak coverage, and channel overload. Detailed definitions of these target categories are provided in Table~\ref{tab:category_description}.

Let each sample $\mathbf{X} \in \mathbb{R}^{m \times l}$ denote the time-series record of a single degraded session, where $m$ is the total number of KPIs aggregated from the PHY, MAC, RLC, and PDCP layers, and $l$ is the temporal length of the monitoring window. Specifically, $\mathbf{X} = [\mathbf{x}_1, \mathbf{x}_2, \ldots, \mathbf{x}_m]^\top$, where each feature vector $\mathbf{x}_i \in \mathbb{R}^{l}$ represents the temporal evolution of the $i$-th KPI over $l$ consecutive intervals.

\begin{table*}
\centering
\caption{Descriptions of Target Classification Categories}
\renewcommand{\arraystretch}{1.1}
\begin{tabular}{|c|c|c|}
\hline
\textbf{ID} & \textbf{Category} & \textbf{Description} \\
\hline
    1 & Uplink Interference & Severe interference detected on the uplink, potentially caused by external sources or frequency overlap. \\
    2 & Uplink Weak Coverage & Low uplink signal strength or quality, often due to poor coverage, penetration loss, or user location. \\
    3 & Downlink Interference & High interference on the downlink, leading to reduced decoding success at the user side. \\
    4 & Downlink Weak Coverage & Weak downlink signal reception caused by distance, obstruction, or poor cell coverage. \\
    5 & Traffic Channel Overload & Overload in the data channel, typically due to high traffic demand and resource exhaustion. \\
    6 & Control Channel Overload & Excessive load on control channels, affecting resource scheduling efficiency. \\
\hline
\end{tabular}
\label{tab:category_description}
\end{table*}

Two types of labeled samples are available for training, reflecting the dual nature of data-and-knowledge-driven supervision:
\begin{itemize}
    \item \textbf{Rule-based labeled samples} $\mathcal{D}_r = \{(\mathbf{X}_r^{(i)}, y_r^{(i)})\}_{i=1}^{N_r}$: rule-based labels are automatically generated at scale by applying domain-specific thresholding rules to KPIs sequences. These labels reflect a data-driven paradigm, leveraging heuristics over large volumes of data but often suffering from noise and limited precision.
    \item \textbf{Expert-based labeled samples} $\mathcal{D}_e =\{(\mathbf{X}_e^{(i)}, y_e^{(i)})\}_{i=1}^{N_e}$: expert-based labels are carefully annotated by experienced communication domain experts based on deep domain knowledge. These labels, though limited in number, embody high-quality human expertise and serve as a knowledge-driven supervisory signal.
\end{itemize}

 Overall, as shown in the Figure~\ref{fig:bigpicture} goal is to learn a classification model $f: \mathbb{R}^{m \times l} \rightarrow {1, 2, \ldots, 6}$ that predicts the most likely root-cause label for a given KPIs time-series input $\mathbf{X}$, effectively leveraging both the abundant but noisy rule-based labels and the limited yet accurate expert annotations. This enables automatic and reliable analysis of root causes behind QoE degradations in mobile communication systems.

\section{Methodology}
\label{sec:proposed method}

\begin{figure}
    \centering
    \includegraphics[width=1\linewidth]{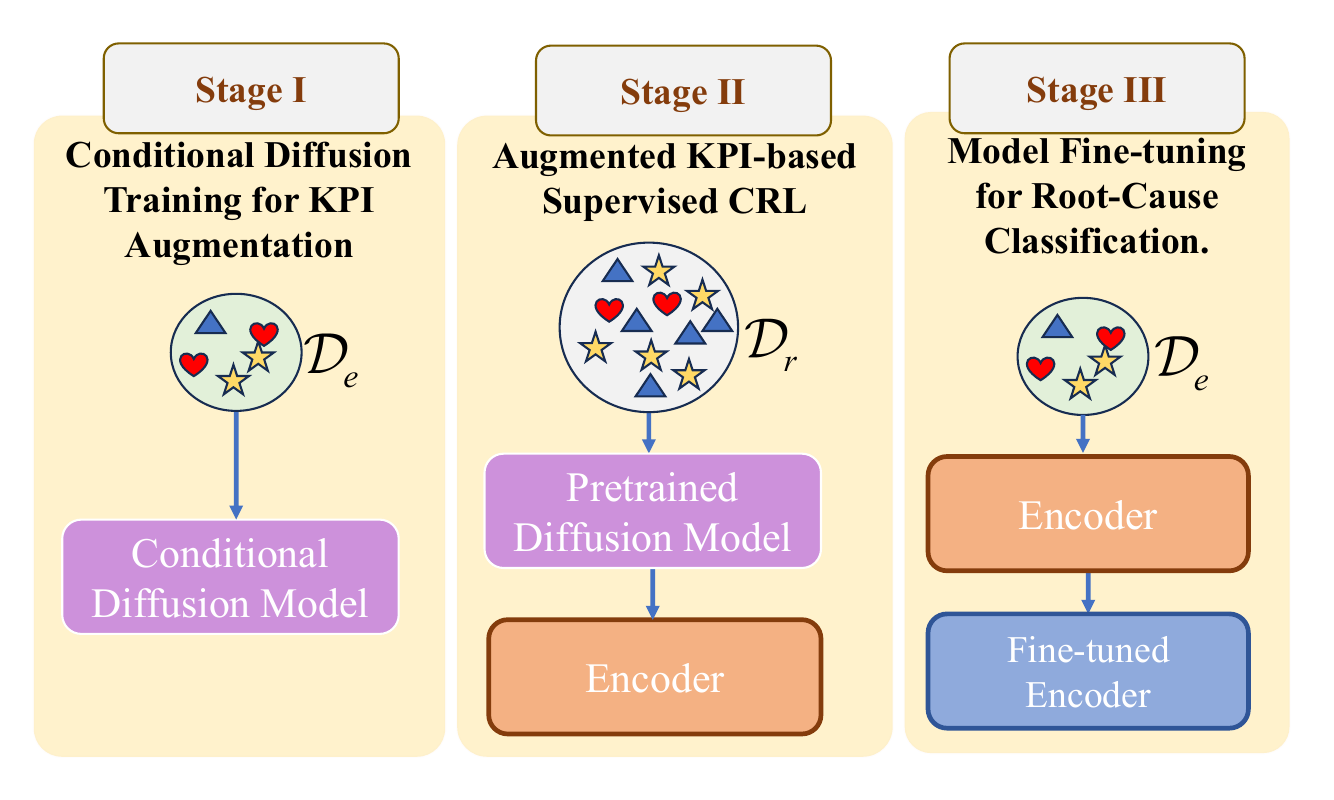}
    \caption{The flowchart of the proposed DK-Root framework.}
    \label{fig: pipeline}
\end{figure}

This paper aims to address the task of classifying the root causes of communication-quality degradation when two sources of supervision are available: a small set of high-fidelity \emph{expert-based} labels and a large set of automatically generated \emph{rule-based} labels whose accuracy is less reliable. As shown in the Figure~\ref{fig: pipeline}, a three-stage learning framework is developed. 

\noindent
\textbf{Stage I-Conditional Diffusion Training for KPIs Augmentation.}  
To leverage expert knowledge for distribution-aware augmentation, we train a label-conditioned diffusion model on the union of expert-labeled samples. Conditioning forces the model to learn class-specific data distributions, and because expert labels are noise-free, they act as an anchor that guides the diffusion process toward the true class manifold.

\noindent
\textbf{Stage II-Augmented KPIs-based Contrastive Representation Learning.}  
To construct a robust representation space under noisy supervision, the diffusion model is frozen and switched to inference mode. For every original sample, we generate two label-consistent views and apply a supervised contrastive loss that pulls embeddings of the same class together and pushes different classes apart, thereby yielding robust and class-discriminative representations without relying on cross-entropy over noisy rule labels.

\noindent
\textbf{Stage III-Model Fine-tuning for Root-Cause Classification.}  
To refine decision boundaries with high-fidelity labels, we attach a lightweight classification head to the encoder and fine-tune the model on the trusted expert-labeled subset, obtaining an effectiveness root-cause classifier.

The remainder of this section provides the implementation details of each stage.

\subsection{Stage I: Conditional Diffusion Training for KPIs Augmentation}
Diffusion models aim to model complex data distributions by gradually perturbing the input data with Gaussian noise and then learning to reverse this corruption process through a sequence of denoising steps. Among them, the Denoising Diffusion Probabilistic Model (DDPM)~\cite{ho2020denoising} has emerged as a foundational framework. 
It defines two stochastic processes: a forward diffusion process that adds Gaussian noise to the input data over multiple steps, progressively destroying information, and a reverse generative process that reconstructs the original data from noisy observations by learning the denoising transitions. The reverse process is parameterized by a neural network trained to predict the added noise at each timestep, which enables sample generation by iteratively denoising pure noise. This simple yet powerful training scheme allows DDPM to approximate highly complex, high-dimensional, and even multi-modal data distributions. 

While unconditional diffusion models are capable of capturing complex data distributions, they lack the ability to guide the generation process toward specific attributes or desired outcomes. To address this limitation, conditional diffusion models have been introduced~\cite{lee2024diffusionnag, dhariwal2021diffusion}. These models incorporate conditioning information (e.g., class labels, auxiliary features, or encoded context) into the reverse denoising steps, thereby enabling the generation of application-specific outputs tailored to task objectives. By injecting relevant external information, conditional diffusion improves controllability and semantic consistency, making it particularly suitable for tasks requiring structured and interpretable data synthesis~\cite{yang2024survey}.

\begin{figure*}
    \centering
    \includegraphics[width=0.7\linewidth]{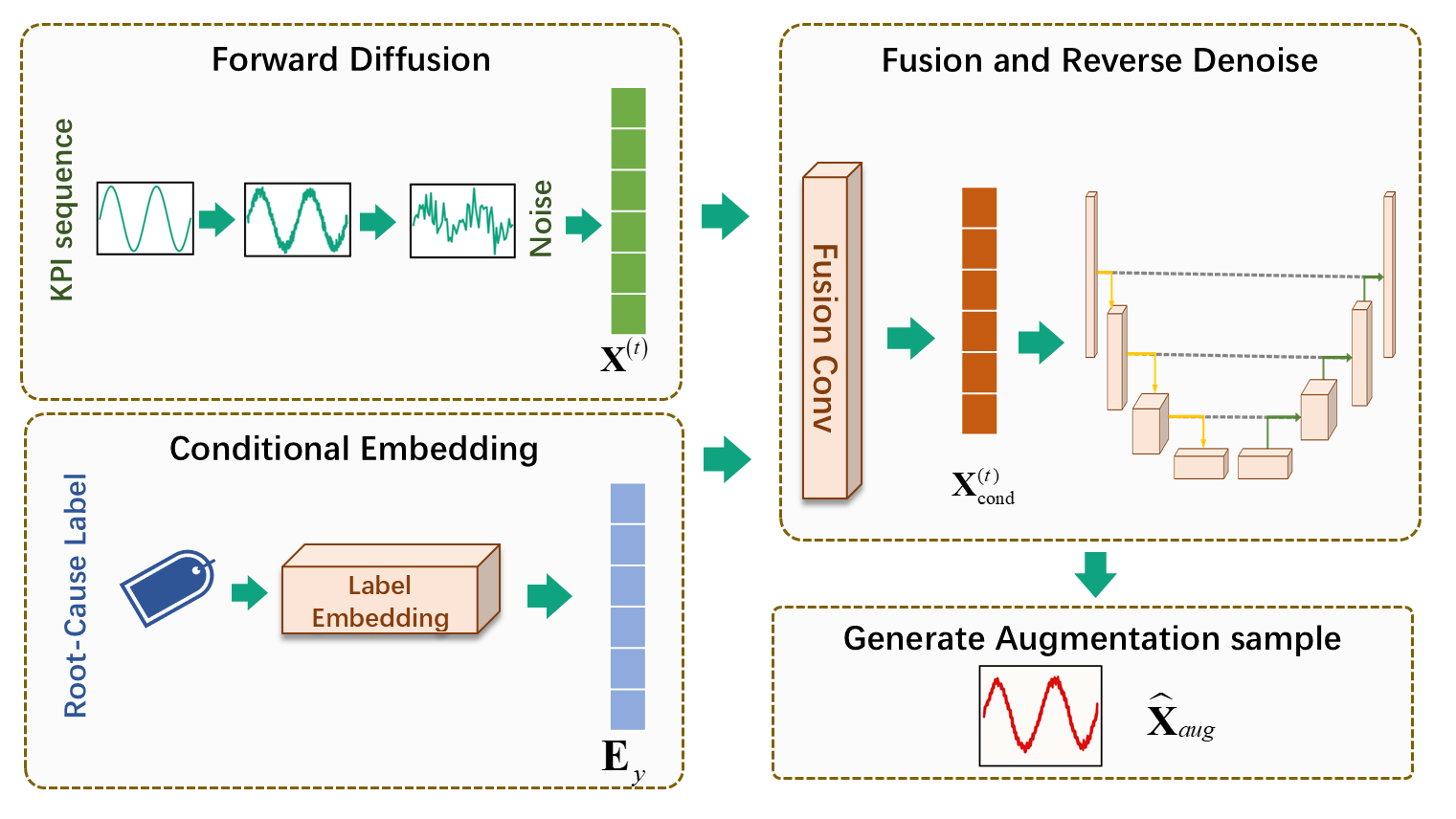}
    \caption{The workflow of the proposed conditional diffusion model.}
    \label{fig: flowchart_cond_diffusion}
\end{figure*}

In our setting, each poor-QoE KPIs sample is denoted as $\mathbf{X} \in \mathbb{R}^{m \times l}$ (equivalently denoted as $\mathbf{X}^{(0)}$, representing the original sample before noise corruption) and the KPIs feature space is governed by a complex distribution $p(\mathbf{X})$, reflecting the intricate impact of diverse degradation causes across network protocol layers.  In this stage, we aim to learn this distribution using the conditional diffusion model illustrated in Figure~\ref{fig: flowchart_cond_diffusion}. Then, by sampling from the learned distribution, the model generates diverse yet semantically aligned augmented samples, which are subsequently used in the contrastive learning stage.

\paragraph{Forward Diffusion}

In the forward diffusion process, We gradually corrupt each clean KPIs time-series sample $\mathbf{X}^{(0)}$ by injecting Gaussian noise over $T$ diffusion steps. This process forms a Markov chain ${\mathbf{X}^{(t)}}_{t=1}^T$, where the distribution of the noisy sample at time step $t$ is given by:
\begin{equation}
q(\mathbf{X}^{(t)} \mid \mathbf{X}^{(0)}) = \mathcal{N}(\mathbf{X}^{(t)};\ \sqrt{\bar{\alpha}_t}, \mathbf{X}^{(0)},\ (1 - \bar{\alpha}_t), \mathbf{I}),
\end{equation}
where $\bar{\alpha}_t = \prod_{s=1}^t \alpha_s$ denotes the cumulative product of noise scaling factors. 

At each time step $t$, the clean input $\mathbf{X}^{(0)}$ is scaled by $\sqrt{\bar{\alpha}_t}$ and perturbed with zero-mean Gaussian noise of variance $(1 - \bar{\alpha}_t)$. This process gradually transforms the structured KPIs sequence into pure noise as $t \rightarrow T$, allowing the model to learn a denoising trajectory during reverse diffusion.

\paragraph{Class-Conditional Embedding and Fusion Module}

Each KPIs sample $\mathbf{X}^{(0)} \in \mathbb{R}^{m \times l}$ is associated with a discrete root-cause label $y$. 
To enable class-aware generative modeling, we incorporate degradation-type labels as conditioning information into the denoising network via a label embedding and early fusion strategy. 

Specifically, we adopt a soft embedding strategy\footnote{In practice, we also embed the diffusion timestep $t$ using a embedding layer and fuse it into the input in a similar manner (details omitted for brevity).}, where each label $y$ is mapped to a continuous vector through a trainable embedding module:
\begin{equation}
\mathbf{e}_y = \mathrm{Embed}_{\text{label}}(y) \in \mathbb{R}^d,
\end{equation}
followed by a linear projection to match the input channel dimension. Then, the projected label vector $\mathbf{e}_y'$ is broadcast along the temporal axis to form a condition tensor:
\begin{equation}
\mathbf{E}_y = \mathbf{e}_y' \otimes \mathbf{1}_l \in \mathbb{R}^{m \times l},
\end{equation}
where $\otimes$ denotes row-wise replication over $l$ time steps.

We concatenate the noisy input $\mathbf{X}^{(t)} \in \mathbb{R}^{m \times l}$ and the label embedding $\mathbf{E}_y$ along the channel dimension:
\begin{equation}
\tilde{\mathbf{X}}^{(t)} = \mathrm{Concat}\left(\mathbf{X}^{(t)},\ \mathbf{E}_y\right) \in \mathbb{R}^{2m \times l}.
\end{equation}
To effectively fuse the label condition with the input sequence, we apply a $1$-D convolutional layer:
\begin{equation}
\mathbf{X}_{\text{cond}}^{(t)} = \mathrm{Conv}(\tilde{\mathbf{X}}^{(t)}) \in \mathbb{R}^{m \times l},
\end{equation}
which projects the concatenated representation back to the original input dimension. 

The condition-aware representation $\mathbf{X}_{\text{cond}}^{(t)}$ is then fed into the U-Net\cite{ronneberger2015u} backbone to predict the noise $\hat{\boldsymbol{\epsilon}}_\theta(\mathbf{X}_{\text{cond}}^{(t)}, t, y)$ at timestep $t$.
During training, diffusion model is optimizes by minimizing the mean squared error (MSE) between the predicted noise and the true noise injected during the forward process:
\begin{equation}
\mathcal{L}_{\text{diff}} = \mathbb{E}_{\mathbf{X}^{(0)}, y, t, \boldsymbol{\epsilon}} \left[ \left\| \hat{\boldsymbol{\epsilon}}_\theta(\mathbf{X}_{\text{cond}}^{(t)}, t, y) - \boldsymbol{\epsilon} \right\|^2 \right],
\end{equation}

This early-stage fusion allows the network to extract class-specific features from the beginning, enabling label-conditioned signal encoding throughout the entire U-Net hierarchy while introduce minimal computational overhead compared to repeated conditioning in intermediate layers.

\paragraph{Reverse Process and Augmentation}
Once trained, the denoising network $\hat{\boldsymbol{\epsilon}}_\theta$ can be leveraged during inference to perform label-guided data augmentation. Unlike standard DDPMs that synthesize inputs from pure noise, our method begins with a real KPI sample $\mathbf{X}^{(0)}$ and generates a semantically aligned perturbation through a controlled diffusion-reversal process.

Specifically, we first apply a forward diffusion step to transform the original sample into its noisy version at timestep $t$, denoted as $\mathbf{X}^{(t)}$. The trained network then predicts the corresponding noise component $\hat{\boldsymbol{\epsilon}}_\theta(\mathbf{X}^{(t)}, t, y)$, conditioned on the degradation type $y$. Then, the augmented view is generated via a single-step reverse update:
\begin{equation}
\hat{\mathbf{X}}_{aug} = \frac{1}{\sqrt{\bar{\alpha}_t}} \left( \mathbf{X}^{(t)} - \sqrt{1 - \bar{\alpha}_t} \cdot \hat{\boldsymbol{\epsilon}}_\theta(\mathbf{X}^{(t)}, t, y) \right).
\end{equation}
where $\bar{\alpha}_t$ denotes the cumulative product of forward noise coefficients up to timestep $t$.

This data augmentation strategy yields semantically aligned views that preserve degradation type while introducing meaningful variations, enhancing representation learning and robustness in downstream tasks.

\subsection{Stage II: Augmented KPIs-based Contrastive Representation Learning}

\begin{figure}
    \centering
    \includegraphics[width=0.85\linewidth]{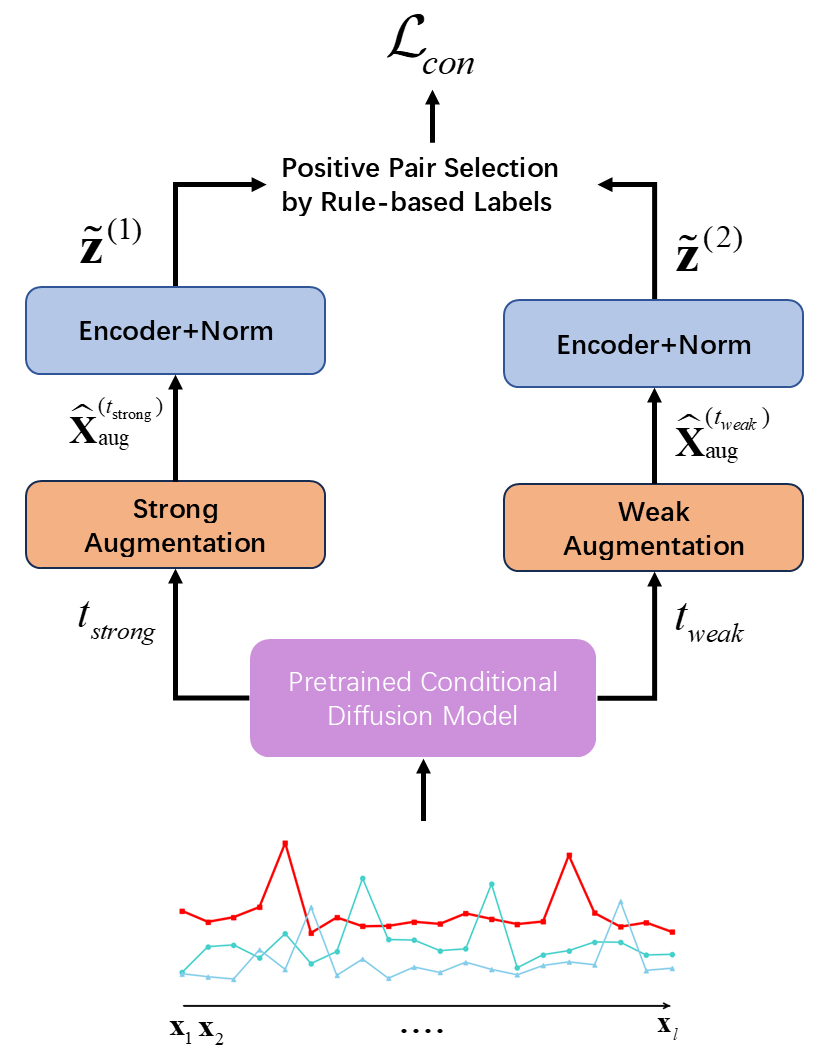}
    \caption{The flowchart of the Stage II.}
    \label{fig:enter-label}
\end{figure}

The core strategy of contrastive learning is to minimize the distance between positive pairs constructed from augmented data. Undoubtedly, data augmentation plays a critical role in contrastive learning. Common augmentation techniques for time series can be broadly categorized into several classes.  
Transformation-based methods~\cite{um2017data, le2016data, goubeaud2021white} apply operations such as jittering, scaling, or magnitude warping directly to raw time-series signals to introduce variability.  Decomposition-based methods~\cite{wen2019robuststl, wen2020fast} decompose time series into subcomponents or segments (e.g., trend, seasonality), augment local timestamps or windows independently, and then reassemble them to construct novel sequences.
In contrast, generative-based methods~\cite{garcia2022improving, yang2023ts, chen2019emotionalgan} aim to capture the underlying data distribution using generative models such as Generative Adversarial Networks (GANs) or Variational Autoencoders (VAEs), from which new synthetic instances can be produced. However, when applied to communication quality degradation sequences, these models still face challenges in reproducing rare but critical anomalies that occur at specific timestamps, distorting critical indicators (e.g., RSRP, SINR) and thereby compromising the integrity of root-cause analysis task.

Traditional forecasting or generative models often make conservative predictions, constraining outputs within the historical value range, and thus struggle to generate low-probability abnormal patterns~\cite{lee2023maat}.  
To address this limitation, diffusion models are increasingly favored
and are incorporated into the proposed framework. They learn to approximate the full data distribution by gradually denoising a Gaussian prior through a reverse diffusion process, enabling the modeling of not only the high-density regions but also the low-probability, anomalous patterns that are critical in communication quality degradation scenarios.  

However, in the task considered in this paper, rule-based labels inevitably contain noise, which can introduce bias into the generated samples if a standard diffusion model is applied directly. While diffusion models are predominantly employed for data generation, their probabilistic formulation also makes them well-suited for data augmentation, where synthetic samples are generated with additional information from the original data. This motivates us to design task-specific augmentation strategies that explicitly account for label imperfections and guide the generation process toward producing semantically consistent, label-aligned variations that enhance the downstream root-cause analysis objective.

\paragraph{Data Augmentation by Pretrained Diffusion Model}
With the conditional diffusion model trained in Stage~I, we perform semantic-preserving data augmentation on the weakly labeled dataset $\{(\mathbf{X}_r^{(i)})\}_{i=1}^{N_r}$, as illustrated in Figure~\ref{fig:enter-label}. 
The objective is to enrich the training set with diverse yet label-consistent variations, thereby improving the performance of representation learning.

Specifically, we introduce two levels of noise injection to create \emph{weak} and \emph{strong} augmentations. Let $T$ denote the total number of diffusion steps in the forward process. 
The augmentation strength is governed by the diffusion time step $t$, which controls the amount of noise added before the denoising process.
For each KPIs sample $\mathbf{X}_r$ in rule-based label dataset, we randomly sample two diffusion steps $t_{\mathrm{weak}}$ and $t_{\mathrm{strong}}$ from non-overlapping subranges of $[1, T]$:
\begin{equation}
t_{\mathrm{weak}} \sim \mathcal{U}(1, \alpha T), 
\quad 
t_{\mathrm{strong}} \sim \mathcal{U}(\beta T, T),
\end{equation}
where $0 < \alpha \leq \beta < 1$ are hyperparameters controlling the relative intensity of weak and strong augmentations. 
Typically, $\alpha$ is set to a small value to ensure minimal semantic distortion, while $\beta$ is set to a larger value to introduce more substantial perturbations.

After injecting noise according to the sampled $t$, the diffusion model performs reverse denoising to generate the augmented samples $\hat{\mathbf{X}}_{\mathrm{aug}}^{(t_{\mathrm{weak}})}$ and $\hat{\mathbf{X}}_{\mathrm{aug}}^{(t_{\mathrm{strong}})}$. 
These augmented views are expected to preserve the intrinsic semantics of the original KPIs sequence while introducing label-consistent variations that sharpen class boundaries across different root-cause categories.

\paragraph{Rule-based Label Aware Representation Learning}
Then, we train the encoder under the supervised contrastive learning paradigm aim to acquire preliminary discriminative ability.

Concretely, the two augmented views are passed through a shared encoder $f_\theta(\cdot)$ to obtain feature representations $\mathbf{z}_1$ and $\mathbf{z}_2$:
\[
\mathbf{z}^{(1)} = f_\theta\!\left(\hat{\mathbf{X}}_{\mathrm{aug}}^{(t_{\mathrm{strong}})}\right), \quad
\mathbf{z}^{(2)} = f_\theta\!\left(\hat{\mathbf{X}}_{\mathrm{aug}}^{(t_{\mathrm{weak}})}\right).
\]
Each representation is then flattened and $\ell_2$-normalized as
\[
\tilde{\mathbf{z}} = \frac{\mathrm{vec}(\mathbf{z})}{\lVert \mathrm{vec}(\mathbf{z}) \rVert_2},
\]
where $\mathrm{vec}(\cdot)$ denotes the vectorization operation.

Let $y_i$ denote the rule-based label assigned to the $i$-th KPIs sample. 
In our supervised contrastive framework, a positive pair is defined as any two samples $(i,j)$ satisfying $y_i = y_j$ and $i \neq j$. 
We use the indicator function $\mathbb{I}[\cdot]$ to represent the positive-pair relation:
\[
M_{ij} = \mathbb{I}[y_i = y_j], \qquad M \in \{0,1\}^{N_r \times N_r}.
\]

Each sample is augmented twice to produce two distinct views, which are encoded and $\ell_2$-normalized to form the contrastive feature set:
\[
\mathcal{C} = \left[ \tilde{\mathbf{z}}^{(1)}_1, \dots, \tilde{\mathbf{z}}^{(1)}_{N_r}, 
\tilde{\mathbf{z}}^{(2)}_1, \dots, \tilde{\mathbf{z}}^{(2)}_{N_r} \right] 
\in \mathbb{R}^{2N_r \times D}.
\]
where $D$ is the dimensionality of the $\ell_2$-normalized embeddings.

The similarity between two feature vectors $\mathbf{a}$ and $\mathbf{c}$ is measured as
\[
s(\mathbf{a},\mathbf{c}) = \frac{\mathbf{a}^\top \mathbf{c}}{\tau},
\]
where $\tau>0$ is the temperature parameter.

Let $s_{ik} = s(\tilde{\mathbf{z}}_i, \tilde{\mathbf{z}}_k)$ 
denote the pairwise similarity between anchor $i$ and contrast feature $k$.
Then, the log-probability is given by the log-softmax over all non-anchor features:
\[
\log p_{ik} = \log \frac{\exp(s_{ik})}{\sum_{k' \neq i} \exp(s_{ik'})}.
\]
Let $P(i) = \{ k \,:\, M_{ik} = 1 \}$ be the positive set for anchor $i$. 
The supervised contrastive loss is defined as
\begin{equation}
\mathcal{L}_{con} = - \frac{1}{2N_r} \sum_{i=1}^{2N_r} 
\frac{1}{|P(i)|} \sum_{k \in P(i)} \log p_{ik}.
\end{equation}
This objective encourages representations of KPIs samples sharing the same rule-based label to be pulled closer, while pushing apart those with different root-cause class, thereby promoting class-discriminative feature learning. The resulting encoder acquires preliminary discriminative ability, which can be further calibrated in a subsequent fine-tuning stage using scarce but high-quality expert annotations.

\subsection{Stage III: Model Fine-tuning for Root-Cause Classification}

The third stage is dedicated to the final degradation classification task. 
While Stage~II produces class-discriminative representations under noisy rule-based supervision, the purpose of this stage is to adapt these representations to high-precision decision boundaries using only clean expert annotations. 
This separation ensures that the learned encoder is first guided by abundant but potentially noisy data, and is then refined using scarce yet reliable labels to maximize classification accuracy.

We use the expert-labeled dataset 
$\mathcal{D}_e = \{(\mathbf{X}_e^{(i)}, y_e^{(i)})\}_{i=1}^{N_e}$ 
for supervised training. 
A lightweight classification head is attached to the encoder obtained from Stage~II, and the entire model is fine-tuned end-to-end to predict degradation categories. 
The classification head outputs class probabilities via a softmax layer:
\begin{equation}
p_c^{(i)} = \frac{\exp(h_c^{(i)})}{\sum_{c'=1}^6 \exp(h_{c'}^{(i)})},
\end{equation}
where $h_c^{(i)}$ is the logit score for class $c$.

We adopt the standard multi-class cross-entropy loss:
\begin{equation}
\mathcal{L}_{sup} = -\frac{1}{N_e} \sum_{i=1}^{N_e} \sum_{c=1}^6 
\mathbb{I}[y_e^{(i)} = c] \, \log p_c^{(i)},
\end{equation}

\section{Experiments}
\label{sec:experiments}

\subsection{Dataset and Setup}

The dataset used in this study is collected by Huawei from real world communication sensors at 5-second intervals, resulting in multivariate time-series samples with a fixed sequence length of 40 (i.e., 200 seconds per sample).
A detailed list of KPIs used in our experiment is provided in Appendix~\ref{appendix}. 
The classification task targets six predefined root-cause categories, as summarized in Table~\ref{tab:category_description}. 
The dataset contains 3,000 samples with rule-based labels (automatically generated by applying structured judging criteria on KPIs) and 185 samples with manually annotated expert labels.

A representative example of a rule-based labeling strategy is illustrated as follows. A sample is assigned the label \textit{Uplink Weak Coverage} if it satisfies all of the following conditions\footnote{We list involved KPIs definitions in Appendix~\ref{appendix} for reference.}:
\begin{itemize}
\item[(1)] \texttt{PDCP\_DL\_LATENCY} $\ge$ 200 ms or \texttt{PDCP\_UL\_LATENCY} $\ge$ 200 ms;
\item[(2)] \texttt{RLC\_DL\_LATENCY} $\ge$ 200 ms or \texttt{RLC\_UL\_LATENCY} $\ge$ 200 ms;
\item[(3)] 
\[
\frac{\texttt{UL\_RLC\_RETX\_SDU}}{\texttt{UL\_RLC\_SDU} + \epsilon} > 0.1, \quad \epsilon = 10^{-5};
\]
\item[(4)] At least one of:
\begin{itemize}
  \item[] \texttt{DL\_RBLER} $\ge$ 0.1,
  \item[] \texttt{UL\_RBLER} $\ge$ 0.1,
  \item[] \texttt{UL\_DTX\_Ratio} $\ge$ 0.2,
  \item[] \texttt{UL\_MAC\_HARQ\_RETX\_MAX} $\ge$ 50, 
  \item[] \texttt{DL\_MAC\_HARQ\_RETX\_MAX} $\ge$ 50;
\end{itemize}
\item[(5)] \texttt{UL\_DMRS\_RSRP\_MIN} $\le$ -125 dBm or \texttt{UL\_SRS\_RSRP} $\le$ -130 dBm.
\end{itemize}


The classification accuracy, defined as the proportion of correctly predicted samples, is used as the main evaluation metric:
\begin{equation}
\text{Accuracy} = \frac{\sum_{i=1}^{N} \mathbb{1}(\hat{y}_i = y_i)}{N}
\end{equation}
where \( \hat{y}_i \) and \( y_i \) denote the predicted and true labels of the \(i\)-th sample, respectively, and \( \mathbb{1}(\cdot) \) is the indicator function.

We conduct three sets of experiments to evaluate the effectiveness of our method. First, we compare against multiple ML and DL baselines to verify its accuracy in root-cause analysis. Second, we validate the proposed conditional diffusion model by examining whether it generates semantically consistent and label-guided augmented samples. Finally, we perform ablation studies and visualizations to assess the impact of contrastive representation learning and superior of proposed framework.

\subsection{Effectiveness of Root-Cause analysis}

\noindent\textbf{Baselines.}  
We compare our method with the following baseline models, including both traditional machine learning (ML) algorithms and state-of-the-art deep learning (DL) methods.\footnote{For traditional ML baselines, we reshape the time-series data into two-dimensional arrays of shape $(n_\text{samples}, n_\text{timesteps} \times n_\text{features})$ before feeding them into the models. For encoder-based DL methods, all models share the same classification head structure, consisting of a fully connected linear layer. For all unsupervised stages or methods in this experiment, rule-based labels are not used during training, and the corresponding samples are treated as unlabeled data. The output of the classification layer is directly used with the cross-entropy loss function, which implicitly applies the softmax operation.}

\begin{itemize}
    \item \textbf{Sliding Window k-Nearest Neighbors (Sliding-KNN):} A non-parametric method based on distance metrics applied to fixed-length subsequences of the time series.
    
    \item \textbf{Statistical Feature-based k-Nearest Neighbors (Stat-KNN):} An extension of KNN that utilizes handcrafted statistical features extracted from the raw time series.

    \item \textbf{Support Vector Machine (SVM):} A classical supervised learning algorithm using the RBF kernel.

    \item \textbf{Time Series Forest (TSF):} An ensemble learning method that trains multiple decision trees on randomly selected intervals of the input time series.

    \item \textbf{Bag-of-Patterns (BOP):} A symbolic representation-based method that first discretizes time series into symbolic words and then applies histogram-based feature extraction for classification.

    \item \textbf{Time-Series representation learning via Temporal and Contextual Contrasting (TS-TCC)~\cite{eldele2021time}}: This DL approach is a semi-supervised framework consisting of two stages. The first stage involves unsupervised contrastive learning with traditional time-series augmentations, while the second stage fine-tunes the model using a small amount of labeled data.

    \item \textbf{Class-Aware Temporal and Contextual Contrasting (CA-TCC)~\cite{eldele2023self}}: This method extends TS-TCC with a four-stage semi-supervised training pipeline. It generates pseudo labels in the third stage using TS-TCC and then refines the model through supervised contrastive learning in the fourth stage.
    
\end{itemize}

\begin{table}
\centering
\caption{Performance comparison (mean accuracy and standard deviation)}
\resizebox{0.45\textwidth}{!}{ 
\begin{tabular}{lcc}
\toprule
\textbf{Method} & \textbf{Mean Accuracy} & \textbf{Standard Deviation} \\
\midrule
SVM             & 0.6113 & 0.0316  \\
Sliding-KNN     & 0.5773 & 0.0435  \\
CA-TCC          & 0.5125 & 0.0435  \\
TS-TCC          & 0.5115 & 0.0651  \\
BOP             & 0.4377 & 0.0158  \\
TSF             & 0.4377 & 0.0310  \\
Stat-KNN        & 0.3547 & 0.0338  \\
\textbf{DK-Root}            & \textbf{0.6830}   & \textbf{0.0370}   \\
\bottomrule
\end{tabular}
\label{tab:baseline_comparison}
}
\end{table}

As shown in the table~\ref{tab:baseline_comparison}, our method achieves the highest mean accuracy of \textbf{0.6830}, outperforming all competing methods across both categories.
Among the traditional ML methods, SVM and Sliding-KNN achieve moderate performance with accuracies of 0.6113 and 0.5773, respectively, demonstrating the effectiveness of simple baselines when applied to time-series classification. However, they are still clearly outperformed by our method, suggesting that incorporating feature representation learning in deep learning can yield substantial benefits over direct classification. 

Compared to recent semi-supervised methods such as TS-TCC (0.5115) and CA-TCC (0.5125), our method shows a significant performance gain. This improvement can be attributed to two main factors: (1) our framework leverages rule-based weak labels during the representation learning stage, providing task-relevant prior knowledge that enhances downstream classification; and (2) the proposed data augmentation strategy is tailored to the root-cause analysis task, as opposed to the generic augmentations (e.g., scaling, jittering) used by TS-TCC and CA-TCC, which may not preserve discriminative temporal patterns.

In terms of robustness, our method exhibits a low standard deviation of \textbf{0.0370}, indicating stable performance across different random seeds. While not the lowest (SVM achieves 0.0316), our model strikes a favorable balance between accuracy and consistency.


\subsection{Effectiveness of Diffusion Augmentation}

This experiment aims to validate the effectiveness of the proposed conditional diffusion-based augmentation strategy. Specifically, we examine whether the generated views preserve the semantic consistency of the class while improving intra-class compactness and inter-class separability under label guidance.

\begin{table*}
\centering
\caption{Comparison of Different Data Augmentation Methods}
\label{tab:augmentation_metrics}
\resizebox{\textwidth}{!}{ 
\begin{tabular}{l|c|ccc|c|cccccc}
\hline
\textbf{Method} 
& \textbf{Avg. Inter-class Dist. $\uparrow$} 
& \textbf{Silhouette $\uparrow$} 
& \textbf{CH $\uparrow$} 
& \textbf{DB $\downarrow$} 
& \textbf{MI $\uparrow$} 
& \textbf{Class1 $\downarrow$} 
& \textbf{Class2 $\downarrow$} 
& \textbf{Class3 $\downarrow$} 
& \textbf{Class4 $\downarrow$} 
& \textbf{Class5 $\downarrow$} 
& \textbf{Class6 $\downarrow$} \\
\hline
Original     & 4.8388 & 0.0185 & 4.3171 & 3.7675 & 0.5430 & 48.944  & 54.8699 & 41.5653 & 45.0379 & 42.8284 & 40.8267 \\
DK-Root(strong)  & \textbf{5.0235} & \textbf{0.0261} & \textbf{4.5135} & \textbf{3.3455} & \textbf{0.5839} & 48.6719 & \textbf{46.1247} & \underline{39.8109} & \textbf{38.1371} & \textbf{37.9744} & 42.0987 \\
DK-Root(weak)   & 4.8141 & \underline{0.0202} & \underline{4.3937} & 3.7287 & \underline{0.5757} & \underline{48.1302} & \underline{53.4976} & 40.3829 & \underline{43.1709} & \underline{41.3713} & \textbf{39.3556} \\
Noise inj        & 4.8221 & 0.0188 & 4.1528 & \underline{3.6324} & 0.4712 & 48.7145 & 57.0828 & 41.3902 & 46.5826 & 42.9834 & 40.8177 \\
Scaling      & \underline{4.9300} & 0.0176 & 4.1618 & \underline{3.6324} & 0.4761 & \textbf{48.1104} & 53.5472 & \textbf{39.7422} & 45.0608 & 42.8211 & \underline{40.5843} \\
\hline
\end{tabular}
}
\end{table*}

\noindent\textbf{Baselines.}  
We compare our method against two commonly used traditional time-series augmentation techniques and the setup is as follows:

\begin{itemize}
    \item \textbf{Proposed method:} For the proposed diffusion-based augmentation, the distribution of $t_{\text{weak}}$ and $t_{\text{strong}}$ are set as follows respectively.
    \begin{align}
        t_{\text{weak}} &\sim \mathcal{U}\left(0,\ \left\lfloor \frac{T}{5} \right\rfloor \right) \\
        t_{\text{strong}} &\sim \mathcal{U}\left(\left\lfloor \frac{T}{5} \right\rfloor,\ \left\lfloor \frac{T}{2} \right\rfloor \right)
    \end{align}

    \item \textbf{Noise Injection}: Gaussian noise is added to each variable based on its temporal statistics. For each sample, we compute the per-channel standard deviation across the time dimension and scale a random Gaussian noise accordingly. The noise is controlled by a scaling factor $\text{ratio} = 0.1$. The operation is defined as:
    \[
        \tilde{x} = x + \text{ratio} \cdot \sigma \cdot \mathcal{N}(0, 1)
    \]
    where $\sigma$ is the per-channel standard deviation computed along the temporal axis.

    \item \textbf{Scaling}: Each variable is perturbed by a randomly sampled scaling factor. The scaling factor for each time point is drawn from a normal distribution $\mathcal{N}(1, \sigma^2)$, where $\sigma = 1.1$, ensuring centered but diverse multiplicative transformations. The operation is applied channel-wise as:
    \[
        \tilde{x}_{i,t} = x_{i,t} \cdot \alpha_t, \quad \alpha_t \sim \mathcal{N}(1, \sigma^2)
    \]
\end{itemize}

\noindent\textbf{Evaluation Metrics.}  
To comprehensively evaluate the quality of the augmented data, we adopt the following quantitative metrics:

\begin{itemize}
    \item \textbf{Average Inter-class Distance}: The mean Euclidean distance between the centers of different classes in the feature space. A larger value indicates better class separability.
    \item \textbf{Silhouette Score~\cite{rousseeuw1987silhouettes}}: Measures how similar a sample is to its own cluster compared to other clusters. Ranges from $-1$ to $1$. Higher scores indicate more coherent clustering.
    \item \textbf{Calinski-Harabasz Index (CH)~\cite{calinski1974dendrite}}: The ratio of between-cluster dispersion to within-cluster dispersion. A higher score indicates better-defined and more compact clusters.
    \item \textbf{Davies-Bouldin Index (DB)~\cite{davies2009cluster}}: Represents the average similarity between each cluster and its most similar one. Lower values imply better inter-cluster distinction.
    \item \textbf{Mutual Information (MI)}: Measures the dependency between features and class labels. A higher mutual information suggests stronger semantic alignment between the learned representation and ground-truth labels.
    \item \textbf{Intra-class Variance}: The average variance of feature vectors within each class. Lower intra-class variance indicates higher compactness and semantic consistency.
\end{itemize}

Experimental results summarized in Table~\ref{tab:augmentation_metrics} clearly demonstrate the superiority of our proposed diffusion strategies over traditional augmentation methods. The \textit{DK-Root(strong)} strategy enhances inter-class separability reflected by the highest average inter-class distance, further supported by clustering metrics(Silhouette, CH, DB), indicating well-separated and compact clusters.
Moreover, \textit{DK-Root(strong)} retains high semantic consistency, as evidenced by the mutual information of 0.5839. It also reduces intra-class variance significantly in categories, demonstrating that generated augmented samples are not only discriminative but also more tightly clustered within the same class. 
On the other hand, \textit{DK-Root(weak)} which is designed for generating minimally perturbed weak views maintains Intra-class variances close to or slightly lower than the original data, demonstrating semantic consistency. Additionally, clustering scores remain comparable to the baseline, confirming that the weak augmentation does not degrade the inherent feature structure.

In contrast, conventional augmentation strategies relatively fail to achieve meaningful improvements. Their inter-class distances remain comparable to the baseline, offering no clear advantage in separability. Furthermore, they suffer from reduced semantic fidelity, as indicated by a drop in mutual information (e.g., 0.4712 for noise, 0.4761 for scaling). Clustering scores also show minimal or no improvement over the original data. These results highlight that random pertutation may disrupt subtle but meaningful patterns (e.g., SINR or RSRP) in communication data, leading to unreliable contrastive learning views.

\begin{figure}
    \centering
    \includegraphics[width = 0.85\linewidth]{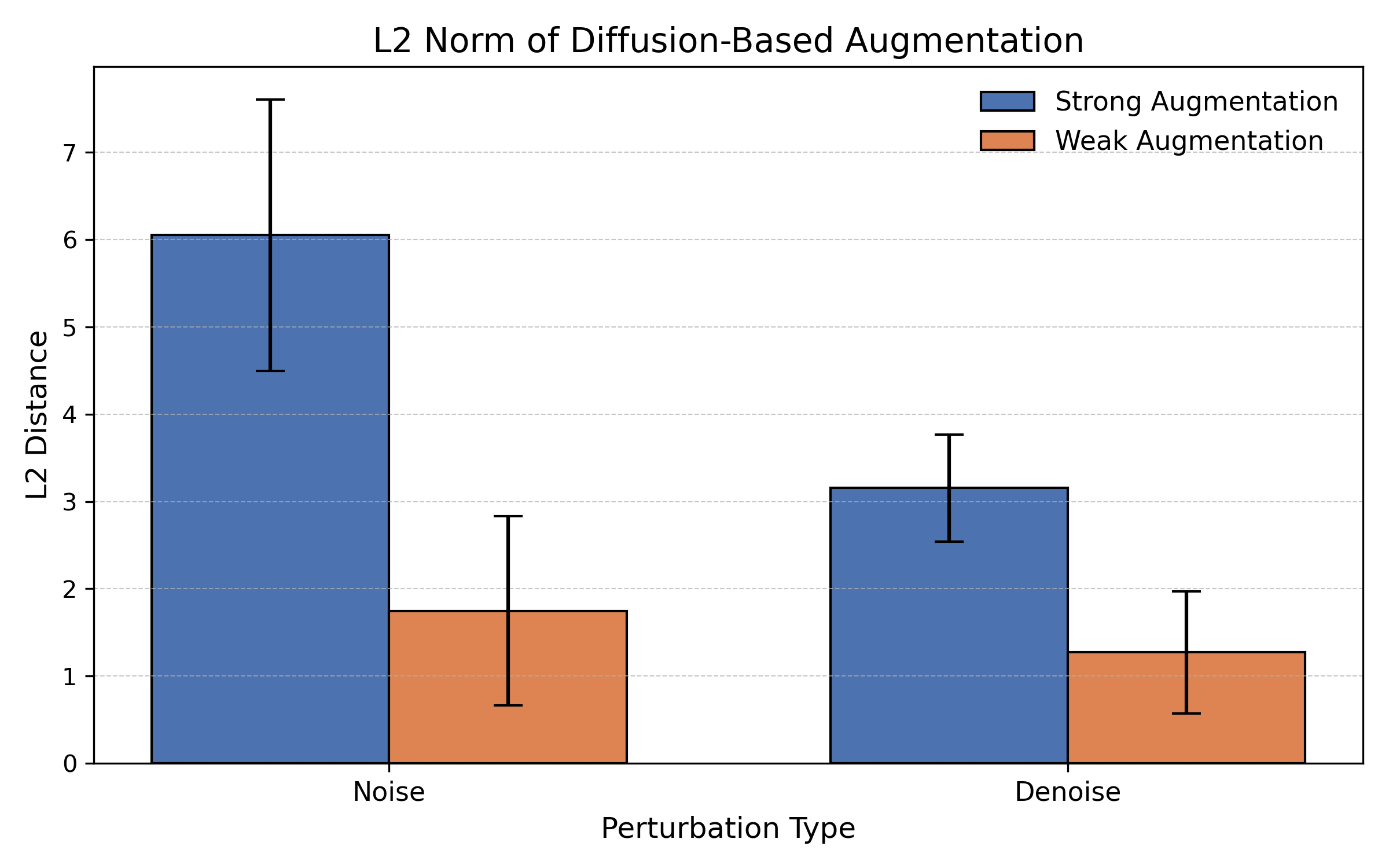}
    \caption{Mean and standard deviation of $L_2$ distances under strong and weak augmentation. “Noise” denotes the distance between the original sample and its noised version before denoising; “Denoise” denotes the distance after reconstruction.}
    \label{fig:two_views}
\end{figure}

Moreover, we further observe the perturbation strength introduced by different time step. Specifically, we compute the $L_2$ distances between the original sample and its augmented versions—both before and after denoising—to quantify the transformation magnitude at different diffusion steps. We report the Noise distance (original vs. noised sample) and Denoise distance (original vs. reconstructed sample).

As shown in Figure~\ref{fig:two_views}, strong augmentation introduces significantly higher perturbation (noise: 6.05, denoise: 3.15) compared to weak augmentation from earlier steps (noise: 1.75, denoise: 1.27). This confirms our design intuition: larger diffusion time steps introduce more sample diversity, while smaller steps yield minimal distortion.
Meanwhile, combining these findings with the quantitative results in Table~\ref{tab:augmentation_metrics}, both of two generated augmentation still preserves class space structure and semantically consistency.

Overall, our results demonstrate that the proposed strong and weak views exhibit controllable perturbation magnitudes while preserving class semantics and clustering structure. This makes them well-suited to serve as augmented views in supervised contrastive learning.

\begin{figure}
    \centering
    \includegraphics[width=0.85\linewidth]{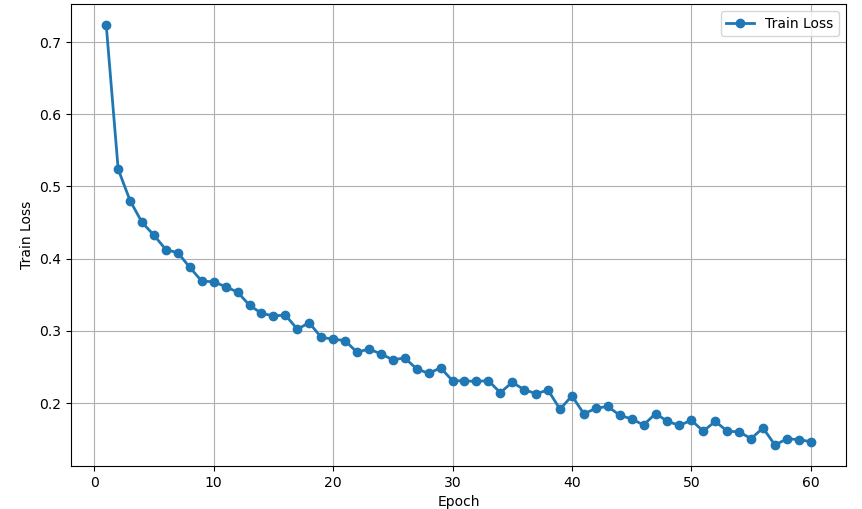}
    \caption{Training loss curve of contrastive learning.}
    \label{fig:loss_curve}
\end{figure}

\begin{figure*}[!t]
    \centering
    \subfloat[Latent space after contrastive learning (Stage II).]{
        \includegraphics[width=0.4\linewidth]{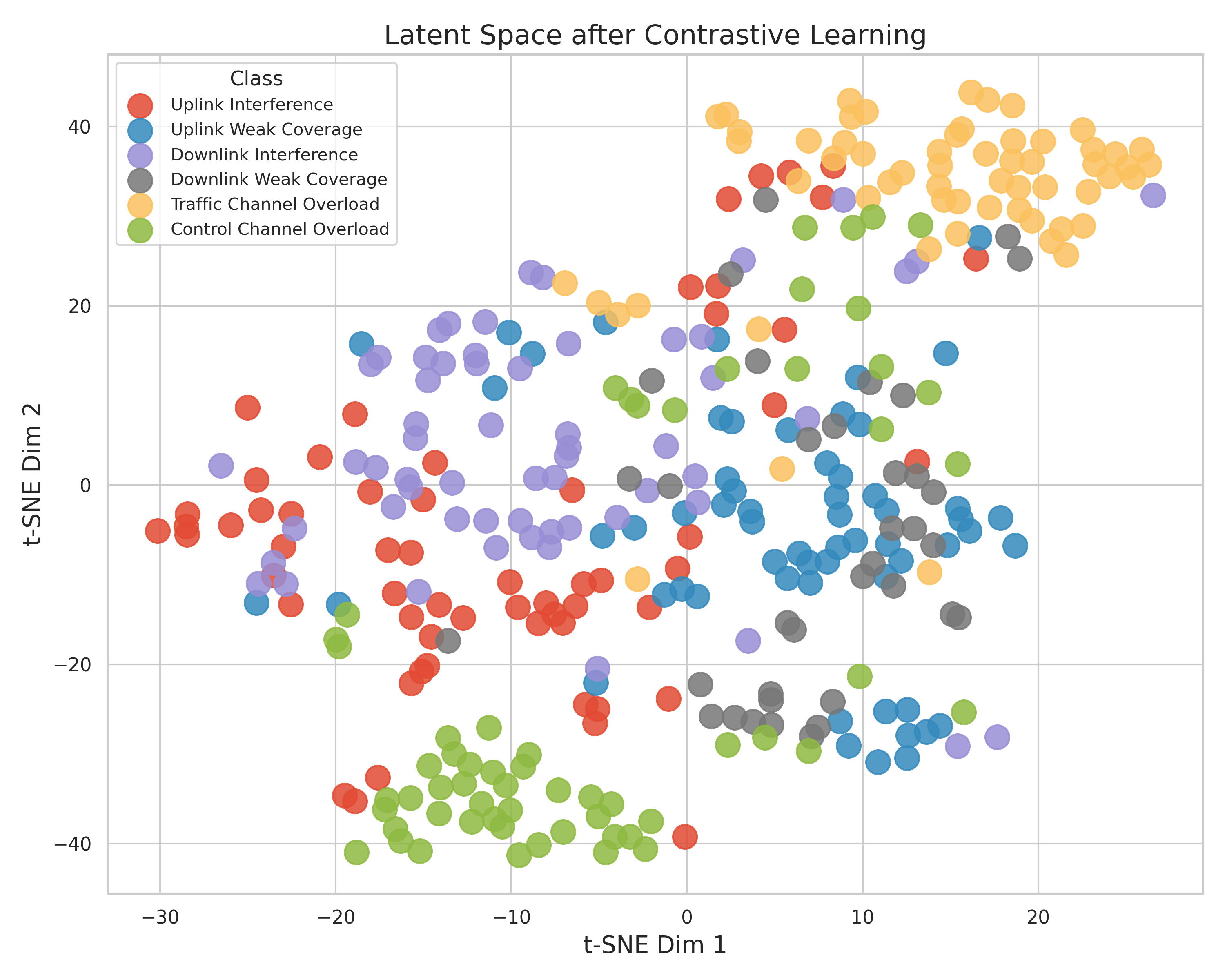}
        \label{fig:tsne_pretrain}
    }
    \hspace{0.05\linewidth}
    \subfloat[Latent space after fine-tuning (Stage III).]{
        \includegraphics[width=0.4\linewidth]{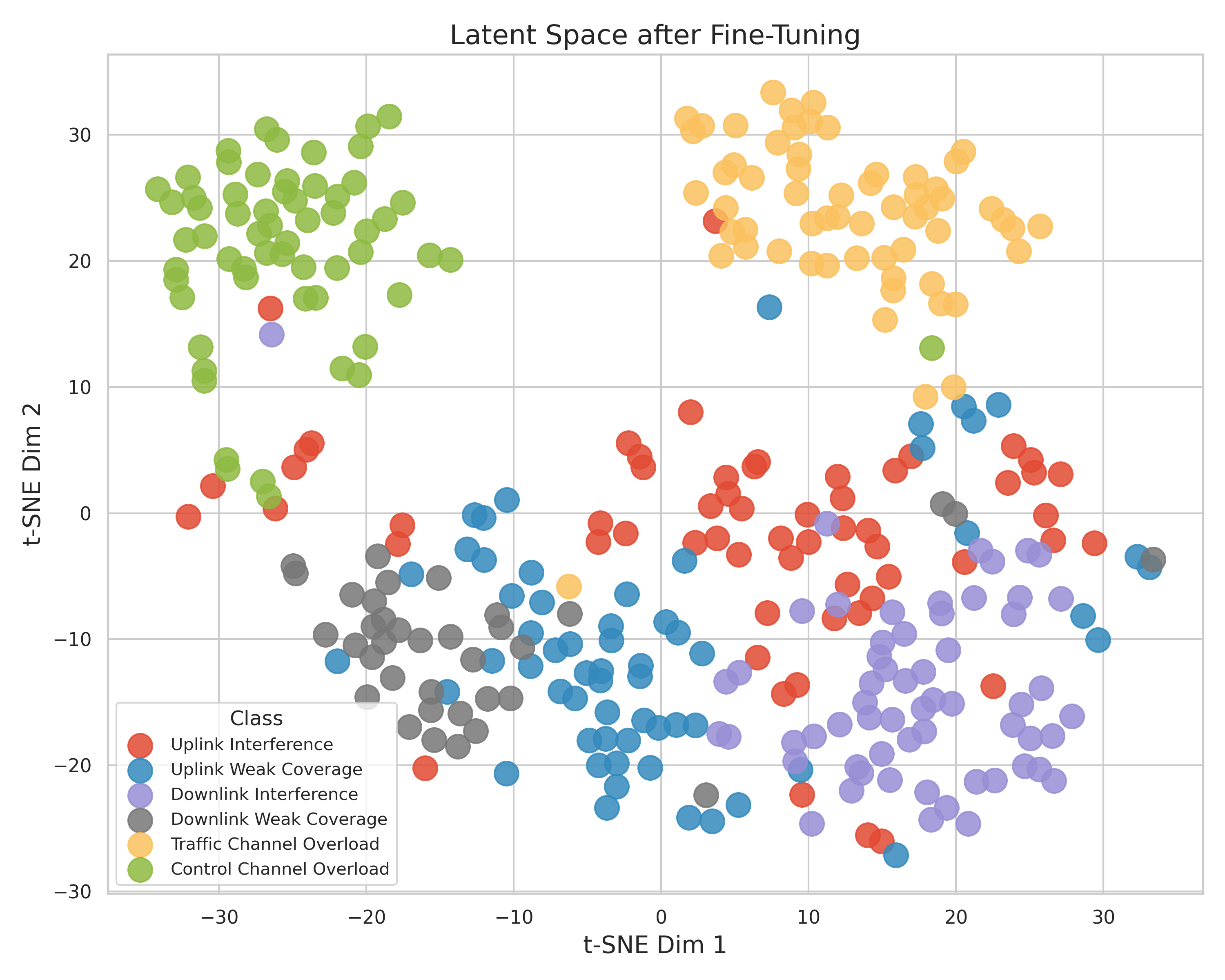}
        \label{fig:tsne_finetune}
    }
    \caption{t-SNE visualizations of the latent space.}
    \label{fig:tsne_comparison}
\end{figure*}

\begin{figure}
    \centering
    \includegraphics[width=0.75\linewidth]{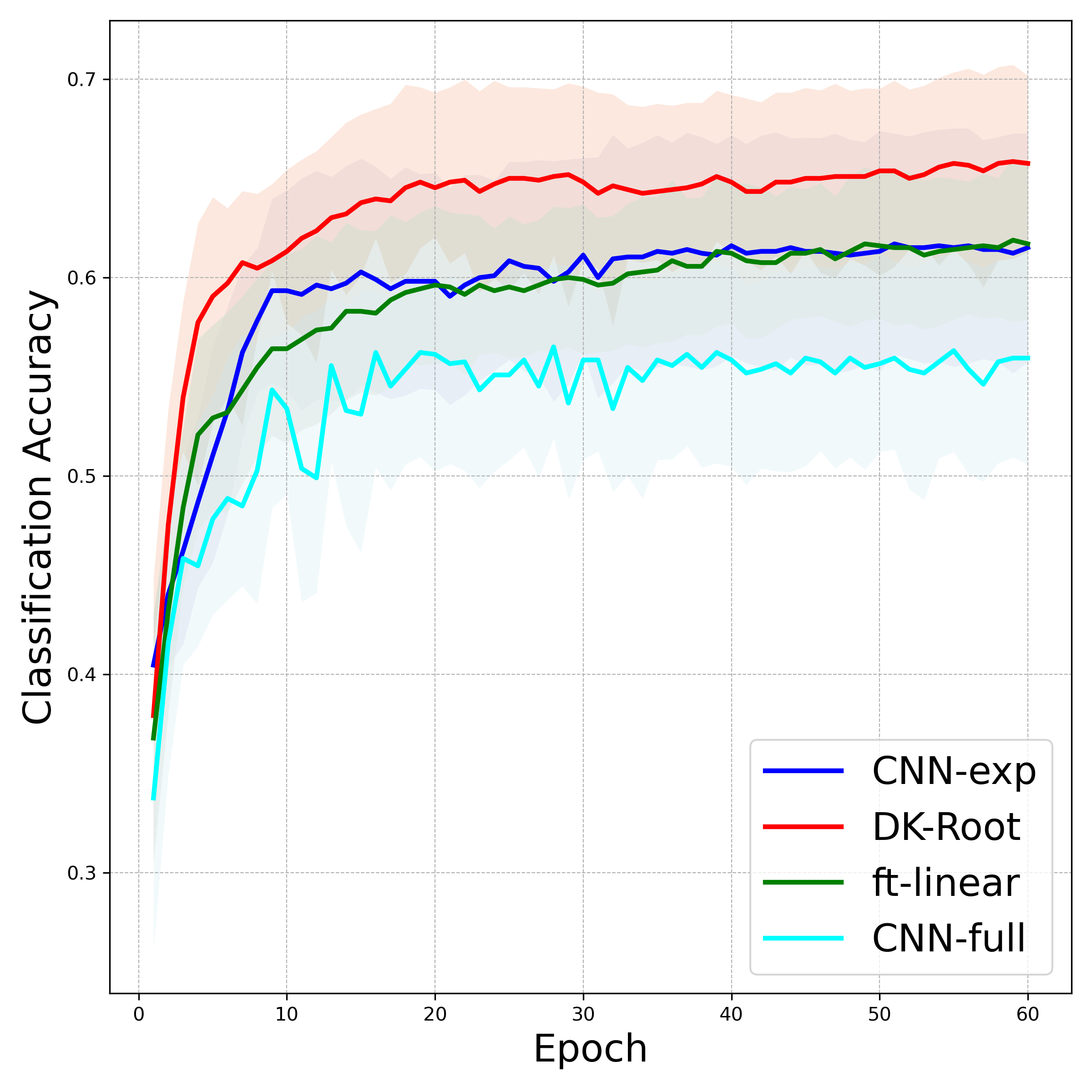}
    \caption{Classification accuracy over epochs for the ablation study.}
    \label{fig:acc_curve}
\end{figure}

\subsection{Priority of pre-train and fine-tuning design}


\noindent\textbf{Effectiveness of Contrastive Learning.}  
In this part, we investigate the effectiveness of the contrastive pretraining stage in our proposed framework. Specifically, we evaluate the structural quality of the learned embeddings using standard clustering metrics and visualize the latent feature space via t-SNE to assess how well the model captures separable representations for downstream classification.

Firstly, the decline curve of the loss function in Figure~\ref{fig:loss_curve} indicates that the model is being properly trained. After approximately 50 epochs, the loss stabilizes around 0.1 with only minor fluctuations, suggesting convergence to a relatively stable representation space.

Next, we visualize the latent feature space using t-SNE in Figure~\ref{fig:tsne_comparison}, where each input sample is passed through the encoder trained at Stage II and Stage III, respectively. The resulting t-SNE plots reveal the structure of the learned feature representations in each case. As shown in Figure~\ref{fig:tsne_comparison}(a), the encoder learns a latent space where samples begin to cluster according to their classes after Stage II (contrastive pretraining), forming rough subspace boundaries. However, noticeable overlaps and mixed regions remain between several classes.
Further, Figure~\ref{fig:tsne_comparison}(b) shows that after Stage III (fine-tuning with expert labels), the latent space becomes significantly more structured and discriminative, with clearer inter-class margins and tighter intra-class clustering. This qualitative improvement is further supported by quantitative clustering metrics: the Silhouette score increases from 0.0565 to 0.2866, the Calinski–Harabasz index rises from 60.36 to 186.40, and the Davies–Bouldin index drops from 7.8031 to 3.4026. These results confirm that contrastive pretraining provides a meaningful initialization, while the subsequent fine-tuning stage substantially enhances the class separability in the learned embedding space.

\vspace{0.5em}
\noindent\textbf{Ablation Study with different strategy.}  
In this part, an ablation study is conducted by evaluating our complete method against three baselines:

\begin{itemize}
\item \textbf{CNN-exp}: A purely supervised CNN\footnote{All CNN-based models share the same network architecture, identical to the encoder used in our proposed method.} trained exclusively on expert-labeled data.
\item \textbf{CNN-full}: A purely supervised CNN trained on the combined dataset (expert and rule-based labels).
\item \textbf{ft-linear}: A classifier that fine-tunes only the classification head.
\end{itemize}

Figure~\ref{fig:acc_curve} summarizes the average accuracy curves for the four compared methods, with mean and standard deviation computed over five random seeds. We first observe that \textit{CNN-exp} achieves a mean accuracy of 0.6481, while \textit{CNN-full}, which incorporates noisy rule-based labels directly into supervised training, sees a notable drop to 0.5981. This nearly 5\% reduction highlights that simply mixing low-quality labels without an intermediate representation learning stage can significantly degrade model performance. In contrast, our method effectively harnesses the complementary strengths of both label types—leveraging rule-based labels for unsupervised representation learning and expert labels for fine-tuning—resulting in enhanced classification accuracy.

To further assess the value of representation learning, we examine the \textit{ft-linear} variant, which freezes the pretrained contrastive backbone and fine-tunes only the classifier head. This approach achieves an accuracy of 0.6349, slightly lower than CNN-exp, but exhibits lower variance (std = 0.0363 vs. 0.0455), suggesting that contrastive pretraining improves the robustness of representations. However, without joint optimization, the model cannot fully exploit these representations for downstream tasks.

Our framework, which jointly fine-tunes both the encoder and classifier, achieves the best performance with a mean accuracy of 0.6830, surpassing \textit{CNN-exp} and \textit{ft-linear} by approximately 3.5\% and 4.8\%, respectively. Moreover, it demonstrates the fastest convergence, exceeding 0.60 accuracy within the first 5 epochs—far earlier than the 20–30 epochs required by baseline methods. These results collectively underscore the importance of joint training and the effectiveness of our proposed three-stage framework.

Overall, these results demonstrate three key insights. First, leveraging noisy rule-based labels through contrastive learning provides robust, noise-tolerant representations that improve both accuracy and convergence speed. Secondly, fine-tuning only the classifier head (ft-linear) is insufficient, and larger performance gains are achieved when both the representation backbone and classifier are jointly optimized. Finally, directly incorporating noisy rule-based labels into supervised training without an intermediate representation learning stage (as in the \textit{CNN-full} baseline) degrades model performance. The above findings indicate the necessity of the model proposed in this article.

\section{Conclusion}
\label{conclusion}

In this paper, we introduce an efficient, scalable semi-supervised solution for classifying root causes of communication quality degradation, with significant practical implications for maintaining mobile network service quality. This framework composes by a three-stage semi-supervised root-cause classification framework that unifies expert annotations and rule-based labels. First, a conditional diffusion model generates diverse, semantically coherent augmentations under label guidance. Next, supervised contrastive learning extracts discriminative features across all samples. Finally, we fine-tune a lightweight classification head on scarce expert-labeled data.
Experimental results on a real-world operator-grade KPIs time-series dataset demonstrate that our approach achieves high accuracy while exhibiting strong stability. Moreover, the experiment results further confirm the rationality and effectiveness of our approach in dual-label scenarios, significantly improving the robustness and accuracy of RCA in communication quality degradation tasks.

However, while our method assumes that rule-based labels are broadly reliable, in scenarios with high label noise, these erroneous signals may be amplified. Future research will focus on robust noise correction by incorporating label-noise modeling or trusted supervision mechanisms to enhance generalization and robustness.

{\appendices

\section{Description of KPIs Used in the Experiments}
\label{appendix}

This appendix summarizes the key performance indicators (KPIs) used in our experiments.
Specifically, Table~\ref{tab:rule_KPI} presents the KPIs employed in the rule-based labeling strategy example,
whereas Table~\ref{tab:selected_features} lists all KPI inputs included in the dataset, together with their detailed descriptions and corresponding protocol-layer mapping

\begin{table}[!htbp]
\centering
\caption{Descriptions of KPIs Used in the Rule-Based Sample.}
\scriptsize  
\label{tab:rule_KPI}
\begin{threeparttable}
\begin{tabular}{@{}p{3.6cm}p{3.7cm}p{0.5cm}@{}}
\hline
\textbf{Feature Name} & \textbf{Description} & \textbf{Layer} \\
\hline
{PDCP\_UL\_LATENCY}        & Average uplink delay on PDCP layer & PDCP \\
{PDCP\_DL\_LATENCY}        & Average downlink delay on PDCP layer & PDCP \\
{RLC\_UL\_LATENCY}         & Average uplink delay on RLC layer & RLC \\
{RLC\_DL\_LATENCY}         & Average downlink delay on RLC layer & RLC \\
{UL\_RLC\_RETX\_SDU}       & Retransmitted SDU segments in UL RLC & RLC \\
{UL\_RLC\_SDU}             & Total transmitted SDUs in UL RLC & RLC \\
{UL\_RBLER}                & Uplink block error rate & PHY \\
{DL\_RBLER}                & Downlink block error rate & PHY \\
{UL\_DTX\_Ratio}           & Ratio of DTX in UL & MAC \\
{UL\_MAC\_HARQ\_RETX\_MAX} & Max HARQ retransmissions in UL & MAC \\
{DL\_MAC\_HARQ\_RETX\_MAX} & Max HARQ retransmissions in DL & MAC \\
{UL\_DMRS\_RSRP\_MIN}      & Min RSRP of UL DMRS signal & PHY \\
{UL\_SRS\_RSRP}            & RSRP of UL SRS signal & PHY \\
\hline
\end{tabular}
\end{threeparttable}
\label{tab:kpi_description}
\end{table}

\begin{table}
\scriptsize
\caption{Descriptions and Protocol Layer Mapping of Selected KPIs.}
\label{tab:selected_features}
\begin{tabular}{@{}p{3.44cm}p{4.05cm}p{0.6cm}@{}}
\hline
\textbf{Feature Name} & \textbf{Description} & \textbf{Layer} \\
\hline
{UL\_DMRS\_SINR} & SINR of the Uplink Demodulation Reference Signal (DMRS), indicating uplink signal quality & PHY \\
{UL\_WB\_PRE\_SINR} & Wideband pre-SINR measured at the uplink side before decoding, reflecting overall uplink signal quality & PHY \\
{UL\_SRS\_RSRP} & Received Signal Received Power (RSRP) of Uplink Sounding Reference Signal (SRS), used for uplink channel quality estimation & PHY \\
{UL\_DMRS\_RSRP\_AVG} & Average RSRP of Uplink DMRS, indicating average received power of uplink reference signal & PHY \\
{SS\_SINR} & SINR of the Synchronization Signal (SS) of the serving cell, representing downlink synchronization quality & PHY \\
{RSRP\_DIFF\_NEIGH\_CNT} & Number of same-frequency neighboring cells whose RSRP differs from the serving cell by more than a specified threshold (Thd1) & PHY \\
{TOP1\_NEIGH\_SSB\_RSRP} & RSRP of the strongest same-frequency neighboring SSB & PHY \\
{SERV\_SSB\_RSRP} & RSRP of the Serving Cell Synchronization Signal Block (SSB) & PHY \\
{SERV\_SSB\_RSRQ} & RSRQ (Reference Signal Received Quality) of the Serving Cell SSB & PHY \\
{CQI\_AVG\_CW0} & Average Channel Quality Indicator (CQI) for codeword 0, used for downlink modulation and coding scheme selection & PHY \\
{DL\_PRB\_UTIL} & Downlink Physical Resource Block (PRB) utilization ratio of the cell & MAC \\
{UL\_PRB\_UTIL} & Uplink Physical Resource Block (PRB) utilization ratio of the cell & MAC \\
{DL\_CCE\_FAIL\_RATE} & Failure rate of downlink Control Channel Element (CCE) allocation & MAC \\
{UL\_CCE\_FAIL\_RATE} & Failure rate of uplink CCE allocation & MAC \\
{CCE\_UTIL\_RATE} & Overall utilization rate of CCEs in the cell & MAC \\
{DL\_CCE\_USAGE\_RATIO} & Proportion of CCEs used for downlink transmission & MAC \\
{COMMON\_CCE\_USAGE} & Proportion of common CCEs shared by control and data channels & MAC \\
{UL\_CCE\_USAGE\_RATIO} & Proportion of CCEs used for uplink transmission & MAC \\
{UL\_SCHED\_FAIL\_RATE\_CCE} & Uplink scheduling failure rate caused by CCE allocation failure & MAC \\
{UL\_SCHED\_FAIL\_CNT\_CCE} & Count of uplink scheduling failures due to insufficient CCE resources & MAC \\
{DL\_SCHED\_FAIL\_RATE\_CCE} & Downlink scheduling failure rate caused by CCE allocation failure & MAC \\
{DL\_SCHED\_FAIL\_CNT\_CCE} & Count of downlink scheduling failures due to CCE shortage & MAC \\
{DL\_CCE\_FAIL\_CNT\_TOTAL} & Total count of downlink CCE allocation failures caused by overall CCE shortage & MAC \\
{DL\_CCE\_FAIL\_RATE\_TOTAL} & Failure rate of downlink CCE allocation due to overall CCE shortage & MAC \\
{DL\_CCE\_FAIL\_CNT\_QUOTA} & Count of downlink CCE allocation failures due to exceeding quota limits & MAC \\
{DL\_CCE\_FAIL\_CNT\_CONFLICT} & Count of downlink CCE allocation failures due to search space conflicts & MAC \\
{UL\_CCE\_FAIL\_CNT\_TOTAL} & Total count of uplink CCE allocation failures caused by overall CCE shortage & MAC \\
{UL\_CCE\_FAIL\_RATE\_TOTAL} & Failure rate of uplink CCE allocation due to overall CCE shortage & MAC \\
{UL\_CCE\_FAIL\_CNT\_CONFLICT} & Count of uplink CCE allocation failures caused by search space conflicts & MAC \\
{DL\_RLC\_TPUT} & Downlink Radio Link Control (RLC) layer throughput in Mbit & RLC \\
{DL\_RLC\_LASTTTI\_RATIO} & Ratio of downlink RLC traffic in the last Transmission Time Interval (TTI) & RLC \\
{UL\_RLC\_TPUT} & Uplink RLC layer throughput in Mbit & RLC \\
{UL\_RLC\_SMALLPKT\_RATIO} & Ratio of small packet traffic in the uplink RLC layer & RLC \\
{PDCP\_DL\_TPUT} & Total downlink PDCP layer throughput (Mbit) & PDCP \\
{PDCP\_DL\_LATENCY} & Average latency of downlink PDCP packets (ms) & PDCP \\
{PDCP\_UL\_TPUT} & Total uplink PDCP layer throughput (Mbit) & PDCP \\
{PDCP\_UL\_LATENCY} & Average latency of uplink PDCP packets (ms) & PDCP \\
\hline
\end{tabular}
\end{table}
\FloatBarrier

}

\bibliographystyle{IEEEtran}
\bibliography{ref}

@article{zhang2024self,
  title={Self-supervised learning for time series analysis: Taxonomy, progress, and prospects},
  author={Zhang, Kexin and Wen, Qingsong and Zhang, Chaoli and Cai, Rongyao and Jin, Ming and Liu, Yong and Zhang, James Y and Liang, Yuxuan and Pang, Guansong and Song, Dongjin and others},
  journal={IEEE transactions on pattern analysis and machine intelligence},
  year={2024},
  publisher={IEEE}
}

@article{eldele2021time,
  title={Time-series representation learning via temporal and contextual contrasting},
  author={Eldele, Emadeldeen and Ragab, Mohamed and Chen, Zhenghua and Wu, Min and Kwoh, Chee Keong and Li, Xiaoli and Guan, Cuntai},
  journal={arXiv preprint arXiv:2106.14112},
  year={2021}
}

@article{eldele2023self,
  title={Self-supervised contrastive representation learning for semi-supervised time-series classification},
  author={Eldele, Emadeldeen and Ragab, Mohamed and Chen, Zhenghua and Wu, Min and Kwoh, Chee-Keong and Li, Xiaoli and Guan, Cuntai},
  journal={IEEE Transactions on Pattern Analysis and Machine Intelligence},
  volume={45},
  number={12},
  pages={15604--15618},
  year={2023},
  publisher={IEEE}
}

@inproceedings{yue2022ts2vec,
  title={Ts2vec: Towards universal representation of time series},
  author={Yue, Zhihan and Wang, Yujing and Duan, Juanyong and Yang, Tianmeng and Huang, Congrui and Tong, Yunhai and Xu, Bixiong},
  booktitle={Proceedings of the AAAI conference on artificial intelligence},
  volume={36},
  number={8},
  pages={8980--8987},
  year={2022}
}

@inproceedings{lee2023maat,
  title={Maat: Performance metric anomaly anticipation for cloud services with conditional diffusion},
  author={Lee, Cheryl and Yang, Tianyi and Chen, Zhuangbin and Su, Yuxin and Lyu, Michael R},
  booktitle={2023 38th IEEE/ACM International Conference on Automated Software Engineering (ASE)},
  pages={116--128},
  year={2023},
  organization={IEEE}
}

@article{wen2020time,
  title={Time series data augmentation for deep learning: A survey},
  author={Wen, Qingsong and Sun, Liang and Yang, Fan and Song, Xiaomin and Gao, Jingkun and Wang, Xue and Xu, Huan},
  journal={arXiv preprint arXiv:2002.12478},
  year={2020}
}

@article{peng2021fault,
  title={Fault feature extractor based on bootstrap your own latent and data augmentation algorithm for unlabeled vibration signals},
  author={Peng, Tengyi and Shen, Changqing and Sun, Shilong and Wang, Dong},
  journal={IEEE Transactions on Industrial Electronics},
  volume={69},
  number={9},
  pages={9547--9555},
  year={2021},
  publisher={IEEE}
}

@inproceedings{luo2023time,
  title={Time series contrastive learning with information-aware augmentations},
  author={Luo, Dongsheng and Cheng, Wei and Wang, Yingheng and Xu, Dongkuan and Ni, Jingchao and Yu, Wenchao and Zhang, Xuchao and Liu, Yanchi and Chen, Yuncong and Chen, Haifeng and others},
  booktitle={Proceedings of the AAAI Conference on Artificial Intelligence},
  volume={37},
  number={4},
  pages={4534--4542},
  year={2023}
}

@inproceedings{um2017data,
  title={Data augmentation of wearable sensor data for parkinson’s disease monitoring using convolutional neural networks},
  author={Um, Terry T and Pfister, Franz MJ and Pichler, Daniel and Endo, Satoshi and Lang, Muriel and Hirche, Sandra and Fietzek, Urban and Kuli{\'c}, Dana},
  booktitle={Proceedings of the 19th ACM international conference on multimodal interaction},
  pages={216--220},
  year={2017}
}

@inproceedings{le2016data,
  title={Data augmentation for time series classification using convolutional neural networks},
  author={Le Guennec, Arthur and Malinowski, Simon and Tavenard, Romain},
  booktitle={ECML/PKDD workshop on advanced analytics and learning on temporal data},
  year={2016}
}

@inproceedings{goubeaud2021white,
  title={White noise windows: data augmentation for time series},
  author={Goubeaud, Maxime and Jou{\ss}en, Philipp and Gmyrek, Nicolla and Ghorban, Farzin and Kummert, Anton},
  booktitle={2021 7th International Conference on Optimization and Applications (ICOA)},
  pages={1--5},
  year={2021},
  organization={IEEE}
}

@article{garcia2022improving,
  title={Improving astronomical time-series classification via data augmentation with generative adversarial networks},
  author={Garc{\'\i}a-Jara, Germ{\'a}n and Protopapas, Pavlos and Est{\'e}vez, Pablo A},
  journal={The Astrophysical Journal},
  volume={935},
  number={1},
  pages={23},
  year={2022},
  publisher={IOP Publishing}
}

@article{yang2023ts,
  title={Ts-gan: Time-series gan for sensor-based health data augmentation},
  author={Yang, Zhenyu and Li, Yantao and Zhou, Gang},
  journal={ACM Transactions on Computing for Healthcare},
  volume={4},
  number={2},
  pages={1--21},
  year={2023},
  publisher={ACM New York, NY}
}

@inproceedings{chen2019emotionalgan,
  title={EmotionalGAN: Generating ECG to enhance emotion state classification},
  author={Chen, Genlang and Zhu, Yi and Hong, Zhiqing and Yang, Zhen},
  booktitle={Proceedings of the 2019 International conference on artificial intelligence and computer science},
  pages={309--313},
  year={2019}
}

@inproceedings{wen2019robuststl,
  title={RobustSTL: A robust seasonal-trend decomposition algorithm for long time series},
  author={Wen, Qingsong and Gao, Jingkun and Song, Xiaomin and Sun, Liang and Xu, Huan and Zhu, Shenghuo},
  booktitle={Proceedings of the AAAI conference on artificial intelligence},
  volume={33},
  number={01},
  pages={5409--5416},
  year={2019}
}

@inproceedings{wen2020fast,
  title={Fast RobustSTL: Efficient and robust seasonal-trend decomposition for time series with complex patterns},
  author={Wen, Qingsong and Zhang, Zhe and Li, Yan and Sun, Liang},
  booktitle={Proceedings of the 26th ACM SIGKDD international conference on knowledge discovery \& data mining},
  pages={2203--2213},
  year={2020}
}

@article{ho2020denoising,
  title={Denoising diffusion probabilistic models},
  author={Ho, Jonathan and Jain, Ajay and Abbeel, Pieter},
  journal={Advances in neural information processing systems},
  volume={33},
  pages={6840--6851},
  year={2020}
}

@inproceedings{lee2024diffusionnag,
  title={DiffusionNAG: Task-guided Neural Architecture Generation with Diffusion Models},
  author={Lee, Hayeon and Ahn, Sohyun and Jo, Jaehyeong and Lee, Seanie and Hwang, Sung Ju},
  booktitle={The Twelfth International Conference on Learning Representations},
  year={2024},
  organization={International Conference on Learning Representations}
}

@article{dhariwal2021diffusion,
  title={Diffusion models beat gans on image synthesis},
  author={Dhariwal, Prafulla and Nichol, Alexander},
  journal={Advances in neural information processing systems},
  volume={34},
  pages={8780--8794},
  year={2021}
}

@article{yang2024survey,
  title={A survey on diffusion models for time series and spatio-temporal data},
  author={Yang, Yiyuan and Jin, Ming and Wen, Haomin and Zhang, Chaoli and Liang, Yuxuan and Ma, Lintao and Wang, Yi and Liu, Chenghao and Yang, Bin and Xu, Zenglin and others},
  journal={arXiv preprint arXiv:2404.18886},
  year={2024}
}

@inproceedings{sun2024spotlight,
  title={SpotLight: Accurate, explainable and efficient anomaly detection for Open RAN},
  author={Sun, Chuanhao and Pawar, Ujjwal and Khoja, Molham and Foukas, Xenofon and Marina, Mahesh K and Radunovic, Bozidar},
  booktitle={Proceedings of the 30th Annual International Conference on Mobile Computing and Networking},
  pages={923--937},
  year={2024}
}

@inproceedings{chawla2020interpretable,
  title={Interpretable unsupervised anomaly detection for RAN cell trace analysis},
  author={Chawla, Ashima and Jacob, Paul and Feghhi, Saman and Rughwani, Devashish and van der Meer, Sven and Fallon, Sheila},
  booktitle={2020 16th International Conference on Network and Service Management (CNSM)},
  pages={1--5},
  year={2020},
  organization={IEEE}
}

@inproceedings{ciocarlie2014feasibility,
  title={On the feasibility of deploying cell anomaly detection in operational cellular networks},
  author={Ciocarlie, Gabriela and Lindqvist, Ulf and Nitz, Kenneth and Nov{\'a}czki, Szabolcs and Sanneck, Henning},
  booktitle={2014 IEEE Network Operations and Management Symposium (NOMS)},
  pages={1--6},
  year={2014},
  organization={IEEE}
}

@inproceedings{moulay2020novel,
  title={A novel methodology for the automated detection and classification of networking anomalies},
  author={Moulay, Mohamed and Leiva, Rafael Garcia and Maroni, Pablo J Rojo and Lazaro, Javier and Mancuso, Vincenzo and Anta, Antonio Fernandez},
  booktitle={IEEE INFOCOM 2020-IEEE Conference on Computer Communications Workshops (INFOCOM WKSHPS)},
  pages={780--786},
  year={2020},
  organization={IEEE}
}

@inproceedings{yuan2020anomaly,
  title={Anomaly detection and root cause analysis enabled by artificial intelligence},
  author={Yuan, Yannan and Yang, Jiaolong and Duan, Ran and Chih-Lin, I and Huang, Jinri},
  booktitle={2020 IEEE Globecom Workshops (GC Wkshps},
  pages={1--6},
  year={2020},
  organization={IEEE}
}

@article{ramirez2023explainable,
  title={Explainable machine learning for performance anomaly detection and classification in mobile networks},
  author={Ram{\'\i}rez, Juan M and D{\'\i}ez, Fernando and Rojo, Pablo and Mancuso, Vincenzo and Fern{\'a}ndez-Anta, Antonio},
  journal={Computer Communications},
  volume={200},
  pages={113--131},
  year={2023},
  publisher={Elsevier}
}

@article{luo2023srcon,
  title={SRCON: A data-driven network performance simulator for real-world wireless networks},
  author={Luo, Zhi-Quan and Zheng, Xi and L{\'o}pez-P{\'e}rez, David and Yan, Qi and Chen, Xin and Wang, Nanbin and Shi, Qingjiang and Chang, Tsung-Hui and Garcia-Rodriguez, Adrian},
  journal={IEEE Communications Magazine},
  volume={61},
  number={6},
  pages={96--102},
  year={2023},
  publisher={IEEE}
}

@article{chen2024overview,
  title={An overview of domain-specific foundation model: key technologies, applications and challenges},
  author={Chen, Haolong and Chen, Hanzhi and Zhao, Zijian and Han, Kaifeng and Zhu, Guangxu and Zhao, Yichen and Du, Ying and Xu, Wei and Shi, Qingjiang},
  journal={arXiv preprint arXiv:2409.04267},
  year={2024}
}

@article{bagnall2017great,
  title={The great time series classification bake off: a review and experimental evaluation of recent algorithmic advances},
  author={Bagnall, Anthony and Lines, Jason and Bostrom, Aaron and Large, James and Keogh, Eamonn},
  journal={Data mining and knowledge discovery},
  volume={31},
  pages={606--660},
  year={2017},
  publisher={Springer}
}

@article{baydogan2015learning,
  title={Learning a symbolic representation for multivariate time series classification},
  author={Baydogan, Mustafa Gokce and Runger, George},
  journal={Data Mining and Knowledge Discovery},
  volume={29},
  pages={400--422},
  year={2015},
  publisher={Springer}
}

@article{ismail2019deep,
  title={Deep learning for time series classification: a review},
  author={Ismail Fawaz, Hassan and Forestier, Germain and Weber, Jonathan and Idoumghar, Lhassane and Muller, Pierre-Alain},
  journal={Data mining and knowledge discovery},
  volume={33},
  number={4},
  pages={917--963},
  year={2019},
  publisher={Springer}
}

@article{fawaz2020deep,
  title={Deep learning for time series classification},
  author={Fawaz, Hassan Ismail},
  journal={arXiv preprint arXiv:2010.00567},
  year={2020}
}

@article{ismail2020inceptiontime,
  title={Inceptiontime: Finding alexnet for time series classification},
  author={Ismail Fawaz, Hassan and Lucas, Benjamin and Forestier, Germain and Pelletier, Charlotte and Schmidt, Daniel F and Weber, Jonathan and Webb, Geoffrey I and Idoumghar, Lhassane and Muller, Pierre-Alain and Petitjean, Fran{\c{c}}ois},
  journal={Data Mining and Knowledge Discovery},
  volume={34},
  number={6},
  pages={1936--1962},
  year={2020},
  publisher={Springer}
}

@article{rousseeuw1987silhouettes,
  title={Silhouettes: a graphical aid to the interpretation and validation of cluster analysis},
  author={Rousseeuw, Peter J},
  journal={Journal of computational and applied mathematics},
  volume={20},
  pages={53--65},
  year={1987},
  publisher={Elsevier}
}

@article{calinski1974dendrite,
  title={A dendrite method for cluster analysis},
  author={Cali{\'n}ski, Tadeusz and Harabasz, Jerzy},
  journal={Communications in Statistics-theory and Methods},
  volume={3},
  number={1},
  pages={1--27},
  year={1974},
  publisher={Taylor \& Francis}
}

@article{davies2009cluster,
  title={A cluster separation measure},
  author={Davies, David L and Bouldin, Donald W},
  journal={IEEE transactions on pattern analysis and machine intelligence},
  number={2},
  pages={224--227},
  year={2009},
  publisher={Ieee}
}

@inproceedings{ronneberger2015u,
  title={U-net: Convolutional networks for biomedical image segmentation},
  author={Ronneberger, Olaf and Fischer, Philipp and Brox, Thomas},
  booktitle={International Conference on Medical image computing and computer-assisted intervention},
  pages={234--241},
  year={2015},
  organization={Springer}
}

\end{document}